**Title Page**

**Title:**

Embedded Chaotic Whale Survival Algorithm for Filter-Wrapper Feature Selection.

**Authors:**

Ritam Guha[1], Manosij Ghosh[1], Shyok Mutsuddi[1], Ram Sarkar[1], Seyedali Mirjalili[2] (corresponding author).

**Affiliations & Addresses:**

[1]Computer Science and Engineering Department, Jadavpur University, Kolkata, India
188, Raja S.C. Mallick Road, Kolkata – 700032, West Bengal, India.

[2]Institute of Integrated and Intelligent Systems, Griffith University, Nathan, Brisbane, QLD 4111, Australia.

**E-mail:** ritamguha16@gmail.com, manosij1996@gmail.com, shyokmutsuddi21@gmail.com, raamsarkar@gmail.com, seyedali.mirjalili@griffithuni.edu.au.

**Corresponding Author Details:**
**Name:** Seyedali Mirjalili
**Email:** seyedali.mirjalili@griffithuni.edu.au

**ORCID:**
Ritam Guha - 0000-0002-1375-777X
Manosij Ghosh - 0000-0003-2954-9876
Ram Sarkar - 0000-0001-8813-4086
Seyedali Mirjalili - 0000-0002-1443-9458

**Abstract:** Classification accuracy provided by a machine learning model depends a lot on the feature set used in the learning process. Feature Selection (FS) is an important and challenging pre-processing technique which helps to identify only the relevant features from a dataset thereby reducing the feature dimension as well as improving the classification accuracy at the same time. The binary version of Whale Optimization Algorithm (WOA) is a popular FS technique which is inspired from the foraging behavior of humpback whales. In this paper, an embedded version of WOA called Embedded Chaotic Whale Survival Algorithm (ECWSA) has been proposed which uses its wrapper process to achieve high classification accuracy and a filter approach to further refine the selected subset with low computation cost. Chaos has been introduced in the ECWSA to guide selection of the type of movement followed by the whales while searching for prey. A fitness-dependent death mechanism has also been introduced in the system of whales which is inspired from the real-life scenario in which whales die if they are unable to catch their prey. The proposed method has been evaluated on 18 well-known UCI datasets and compared with its predecessors as well as some other popular FS methods.

# Embedded Chaotic Whale Survival Algorithm for Filter-Wrapper Feature Selection

**Abstract:** Classification accuracy provided by a machine learning model depends a lot on the feature set used in the learning process. Feature Selection (FS) is an important and challenging pre-processing technique which helps to identify only the relevant features from a dataset thereby reducing the feature dimension as well as improving the classification accuracy at the same time. The binary version of Whale Optimization Algorithm (WOA) is a popular FS technique which is inspired from the foraging behavior of humpback whales. In this paper, an embedded version of WOA called Embedded Chaotic Whale Survival Algorithm (ECWSA) has been proposed which uses its wrapper process to achieve high classification accuracy and a filter approach to further refine the selected subset with low computation cost. Chaos has been introduced in the ECWSA to guide selection of the type of movement followed by the whales while searching for prey. A fitness-dependent death mechanism has also been introduced in the system of whales which is inspired from the real-life scenario in which whales die if they are unable to catch their prey. The proposed method has been evaluated on 18 well-known UCI datasets and compared with its predecessors as well as some other popular FS methods.

## 1. Introduction

With the introduction of various datasets in this digitally advanced era, data mining [1] has become one of the most interesting and challenging techniques to convert the available data into useful information. As the dimension of datasets keeps growing, data mining models face increasing problems due to inclusion of many redundant features in the datasets. To counter this problem, a pre-processing technique called Feature Selection (FS) has gained popularity in the recent years [2]. FS [3] is the process of finding a subset of important features from a dataset which contains relevant as well as redundant features. For an $n$ dimensional feature set, there are $2^n$ possible combinations (feature subsets) which makes FS an NP hard problem. Hence various machine learning (ML) models are employed to solve the FS problem within a reasonable time frame.

The models are broadly classified into two categories: wrapper [4]–[8] and filter [9]–[14]. Wrapper-based models use learning algorithms (e.g. classifiers) to evaluate the resultant subset of features whereas the filter-based approaches use intrinsic properties of features to evaluate their candidate subsets. Wrapper models require significantly high time to evaluate their candidates but they are able to produce a more dominant subset of features whereas filter methods take less time for evaluation but the quality of the generated subsets gets compromised. The recent trend is to take advantages of both filter and wrapper approaches to form more robust models which are known as hybrid or embedded models [15]–[20].

One of the most recent additions to the pool of wrapper methods is Whale Optimization Algorithm (WOA) [21]. WOA replicates the movement of humpback whales while searching for prey to perform FS. Over the years, many variants of WOA have been proposed in the literature [22]–[26] but no embedded version of WOA has been proposed till date to the best of our knowledge. Moreover, some real-life attributes of whales are missing in WOA. For example, all whales in a group are different and hunt in a slightly different manner. Hence, some whales in the population, who are unable to hunt, will eventually die before the rest following the survival of the fittest mechanism. On the other hand, the preys are not located in fixed positions, rather they also move. So, when the whales reach the position of the prey recorded in the past, they may have moved from their previous positions. In order to mimic these real-life scenarios, a new embedded version of WOA has been modelled named Embedded Chaotic Whale Survival Algorithm (ECWSA) with inclusion of death and local search technique. Real-life whales use two procedures while searching their prey – shrinking encircling and spiral motion. In WOA, the whales select one of these techniques depending on a random number. In order to bring a systematic change to this random number, chaotic maps are used to guide the selection of the search procedure.

The key factor in case of any metaheuristic is achieving a proper trade-off between exploitation and exploration. In pursuit of this trade-off, researchers propose different metaheuristic algorithms very often. But some of the most popular metaheuristic algorithms suffer from various drawbacks. Particle Swarm Optimization (PSO), proposed in 1995 by Kennedy [27], is one of the most popular swarm-based metaheuristic algorithms. Although PSO is armed with extensive local search capability, it lacks in exploration ability. In many cases, PSO has the tendency to converge to a local optimum [28]. Another frequently used metaheuristic FS algorithm is Gravitational Search Algorithm (GSA) [29]. In GSA, the candidate solutions (a.k.a masses) attract each other according to the law of gravitation to form better solutions over the iterations. But GSA suffers from the problem of premature convergence. If any intermediate solution in GSA possesses high fitness value, it produces a large force of attraction resulting into faster convergence. Hence, the recent trend is to propose a hybrid of multiple metaheuristic algorithms so that good sides of each algorithm can be used in order to overcome the limitations of the individuals [30]–[33]. Although these hybrid algorithms perform better, they need a proper configuration of communication among the algorithms forming the hybrid models. Sometimes, these configurations are difficult to tune, and non-standard tuning approach fails to take the advantages of candidate solutions. Such limitations are addressed in this work.

Proposed ECWSA is rich in both exploration and exploitation. To be more specific, death of the whales in ECWSA helps in faster convergence. Convergence is very important for any FS algorithm, but sometimes it may lead to pre-mature convergences as well which are undesirable. In order to avoid this phenomenon, chaos is introduced which brings a flavor of restricted randomness in the search which increases exploration. Local search using Minimum Redundancy Maximum Relevance (mRMR) helps us to prune the feature sets using properties of the features. This enhances the algorithm's ability to remove the unnecessary features without much computational costs. These new features allow for a more extensive exploration phase while simultaneously avoiding premature convergence of the solutions. Thus, it can be observed that ECWSA embodies a good combination of exploration and exploitation, fast but without pre-mature convergence along with great local search capability.

The main contributions of the proposed model are as follows:

- mRMR based filter method is used to perform local search. This allows the whales to get to the exact locations of the preys.
- The concept of chaos is introduced to guide the whales in selection of type of movement. This helps to better the search capability of the whales.
- Faster convergence is achieved by the introduction of death in the group of whales. This resembles more closely the real-life scenario in which only fitter whales survive while other whales die due to undernutrition.
- The proposed algorithm has been tested over 18 well-known UCI datasets to prove its applicability and usefulness. It has been additionally applied over 7 Microarray datasets to evaluate the robustness of the algorithm.

The rest of the paper is organized in 5 sections. Section 2 gives a brief description of the related works performed in the same domain. The proposed method ECWSA is described in detail in Section 3. The experimental outcomes, their comparisons, stability checking, and convergence related details are provided in Section 4. Section 5 concludes our proposed work and provides a broad outline of the possible future works.

## 2. Related Work

Due to increasing popularity of FS as an effective pre-processing technique, various researchers have used the concept of meta-heuristics to solve this challenging problem. FS can be viewed as an optimization problem to find an optimal subset of features subject to some constraints (e.g. maximum allowable iterations). Hence, recently people are applying popular optimization algorithms to FS problems and vice versa.

WOA has been basically proposed to solve optimization problems in 2016 by Mirjalili and Lewis [21]. The method has been tested on 29 mathematical functions and 6 structural design problems (*namely* design of a welded beam, design of a tension/compression spring, design of a 25-bar truss, design of a pressure vessel, design of a 15-bar truss, and design of a 52-bar truss de- sign) which concludes its competitiveness with other meta-heuristic and conventional optimizers. The optimization approach adapted by WOA has been modified to solve FS problems in [23] where Mirjalili et al. proposed a few binary variants of WOA. The first two variant uses Roulette wheel and Tournament selection in the search process which are known as WOA-R and WOA-T respectively. The second variant uses crossover and mutation operators to improve the exploitation of basic WOA and it is known as WOA-CM. These FS approaches have been tested over 18 well-known UCI datasets which has revealed that all the variants of WOA are able to achieve better results than some popular FS approaches like Genetic Algorithm (GA) [5], [34], [35],PSO [36]–[38] , Ant Lion Optimizer (ALO) [39], [40] etc. Another FS approach using the concepts of WOA has been proposed in [24] by Sharawi et al.

Mirjalili et al. have proposed a hybridized version of WOA in [22]. In this paper, Simulated Annealing (SA) has been used to enhance exploitation by performing search around the most promising regions located by WOA. Mainly two hybrid variants are proposed: low- level teamwork hybrid model (LTH), and high-level relay hybrid model (HRH). In the first model, SA is used as a local search technique in order to exploit the selected search agents. The second hybrid model uses SA to search the neighborhood of the best solution found after each iteration. Though called a hybrid it should be noted that the model is a wrapper based.

Apart from FS, WOA has been used to solve many other optimization problems. Aljarah et al. have performed neural network parameter optimization using WOA in [41] where search agent of WOA represents a candidate neural network of Multi-Layer Perceptron (MLP). The objective of the optimization is to find optimal values for weights and biases of the neural network and thereby reducing the mean square error (MSE) present in the values predicted by the candidate neural networks. In [42], Oliva et al. have presented a chaotic version of WOA named Chaotic WOA (CWOA) which is used to optimize the parameters of the photovoltaic cells and panels. The chaotic maps help CWOA to compute and automatically adapt the internal parameters of the optimization algorithm. Prakash et al. used WOA [43] to optimize sizing and placement of capacitors in a typical radial distribution system. Operating cost reduction and power loss minimization are considered to be the objectives of the approach. Kaveh et al. have used a modified version of WOA called Enhanced WOA (EWOA) in [44] to optimize sizing of truss and frame structures. Wang et al., in [45], have modified WOA to solve multiple objectives used for wind speed forecasting and named the updated version as Multi Objective WOA (MOWOA). High accuracy and stability are used as the objectives for MOWOA.

A chaotic approach is implanted in ECWSA in order to guide the whale movements. Chaos is a well-known approach to bring randomness in deterministic dynamic system. Zawbaa et al. have introduced chaos in ALO in [46] and applied it to perform FS which has significantly improved the trade-off between exploration and exploitation of ALO. Mafarja et al. in [47] incorporated chaotic maps in Salp Swarm Algorithm (SSA) to perform FS. A chaotic version of PSO has been introduced in the FS domain by Yang et al. in [48]. Apart from these, chaos is also implemented in various other popular FS algorithms like Dragonfly algorithm [49], Crow Search algorithm [50] etc. There are some instances of using chaos alongside WOA as well [51]–[55]. So, it can be seen that chaos is a popular and well-accepted approach in the domain of FS to bring balance between exploration and exploitation.

Sayed et al. have proposed Chaotic WOA (CWOA) in [51]. CWOA uses chaos to guide the movement of every random parameter present in WOA. The authors have tested with 10 chaotic maps and have found that the circular chaotic function works best for the situation. Each time, the chaotic maps have been initialized with 0.7. Restricting the randomness of every random parameter through chaos may end up diminishing the stochastic capabilities of WOA. Moreover, same initial value for every chaotic function used for different parameters guide their values in a similar way which further reduces the randomness of the algorithm. Instead of using discrete time chaotic systems, the authors, in [52], have made an attempt to utilize continuous time chaos to improve the performance of WOA. Based on the results and analysis, they have found that real time chaotic systems can improve the quality of the solutions for multidimensional problems and are able to provide faster convergence. In [53], the authors have used logistic chaotic function to speed up the convergence of WOA. The chaotic WOA is then used to optimize penalty parameter and kernel parameter of kernel Extreme Learning Machine (KELM) which is used to classify mammograms for breast cancer identification. A novel chaotic multi-swarm WOA approach has been proposed in [54] which is used to perform parameter optimization and FS for a SVM-based model. The entire model, called CMWOAFS-SVM, has outperformed other SVM models associated with GA, PSO, basic WOA and Bacterial Foraging Optimization (BFO) algorithms. CWOA has been applied to stability constrained Optimal Power Flow (OPF) problem in [55]. The results indicate that CWOA is able to provide high convergence, stability, better solutions even in OPF domain. So, it is apparent that chaos can help WOA provide faster convergence rates and better solutions. But, again too much use of chaos may result into premature convergence and reduction in stochastic capabilities. In order to use chaos in an effective way, ECWSA uses chaotic map only for one of the most important parameters in WOA instead of applying it for every random parameter.

Apart from the above-mentioned works, a wide range of recent metaheuristics have been employed to solve feature selection problems [56]–[62] as well as other problems [63]–[67]. Some recently proposed hybrid algorithms applied to FS are hybrid BALO [68], hybrid Grey Wolf Optimizer [69], [70], hybrid ACO [18], hybrid GA [71], [72] etc. Thus, the increasing popularity of FS and the application of metaheuristic algorithms in this domain is clearly visible. It can be also observed that WOA is one of the most popular metaheuristics that has been used in wide range of applications in the previous years including FS but no significant effort has been made to develop an embedded version of the same. WOA has no built-in structure to amplify its convergence and balance exploitation and exploration. This fact motivates us to combine the power of wrapper version of WOA with the concepts of chaos, death and filter-based local search. This combination is able to achieve good classification performance with a competent dimensionally reduction ability. Moreover, the combination with a filter method improves the performance without considerable increment in computational complexity.

## 3. Proposed Methodology

Our proposed technique called ECWSA is a modification over a recently developed FS approach named WOA. ECWSA incorporates several things into WOA to improve the performance of the same. It allows the whales to adapt to tougher conditions with passing times (here iterations). This use of harshness allows the fittest whales to survive allowing for a smoother convergence. This high rate of convergence may cause premature convergence (inability to leave a local optima). To counter this, mRMR based filter approach is used to introduce diversity. The utilities of this approach are two-fold, one being that it can help to introduce further exploration and secondly it helps to include data intrinsic properties into selection of feature subset. Section 3.1 provides a detailed description of the proposed method while section 3.2 presents time complexity analysis of the model. It is to be noted that the proposed model is referred to as proposed model, method, version, approach interchangeably in the manuscript but they refer to the same.

### 3.1 ECWSA

Humpback whales hunt krill or small fish in groups where they encircle their prey and trap them in nets of bubbles. The foraging behavior is used for FS by considering each whale as a feature subset. The whales are represented as binary strings ($\{x_1, x_2, \ldots, x_i, \ldots, x_n\}$ of length equal to that of number of features - $n$) where '1' implies that the feature is selected in the subset and '0' otherwise. Each whale is represented by $\vec{X}(t)$. Each whale moves either according to the position of the prey or in search of prey. In ECWSA, the whales try to achieve a balance of exploration and exploitation in their movements. This balance in whales is critical to avoid pitfalls like that of getting stuck in a local optimum or failing to explore the search space properly. Exploitation can be referred to as finding a better solution from the existing explored search space. Exploration deals with moving towards unexplored parts of the search space. Exploration and exploitation require differing movements during the forage for food.

### 3.1.1 Exploitation Stage

In swarm movements of the humpback whales (denoted by $\vec{X}(t)$ in equation 1) the position of prey is encircled by the whales. The position of the prey would be the best binary substring possible. so, in our application the position of the prey can be assigned to the best available subset of features found till that time. Therefore, the position of the best whale ($\vec{X}^*(t)$) found till that time ($t$) is the prey's position towards which other whales try to move to. The movement of whales towards the prey is an exploitation of the search space. This motion towards the best whale is of two kinds *namely* spiral motion or shrinking encircling. The spiral motion occurs according to equation 2. The variable $l$ is random vector of size $n$ in the range of $[-1, 1]$. The whale takes a spiral path towards the best whale ($\vec{X}^*(t)$). The value of $b$ determines the kind of shape the logarithmic spiral has.

$$\vec{X}(t) = \{x_1, x_2, \ldots, x_i, \ldots, x_n\} \text{ at time } t \tag{1}$$

$$\vec{X}(t+1) = |\vec{X}^*(t) - \vec{X}(t)| \cdot e^{bl} \cdot \cos(2\Pi l) + \vec{X}^*(t) \tag{2}$$

Exploitation is also undertaken by using shrinking encircling movement towards the best whale. In this method, the positions of the whales are updated using equation 3.

$$\vec{X}(t+1) = \vec{X}^*(t) - \vec{A}.\vec{D} \tag{3}$$

Here, $\vec{A}$ represents a vector of size $n$ calculated using equation 4 and $\vec{D}$ is the modified distance between the prey and a whale computed using equation 5. The value of $a$ in equation 4 is computed using equation 7 for each iteration. The value of $a$ is decreased from 2 to 0 over the iterations. $\vec{r}$ in equations 4 and 6 is an $n$-dimensional random vector.

$$\vec{A} = 2a.\vec{r} - a \tag{4}$$

$$\vec{D} = |C.\vec{X}^*(t) - \vec{X}(t)| \tag{5}$$

$$\vec{C} = 2 * \vec{r} \tag{6}$$

$$a = 2 - t\frac{2}{maxIter} \tag{7}$$

### 3.1.2 Exploration Stage

Equation 3 lets the whales move closer to the prey which is basically the best whale found so far. On the other hand, if some random whale is chosen to represent the prey, it will lead to the exploration of the search space. Thus, shrinking encircling can lead to both exploitation and exploration. In order to provide exploration to the system of whales, the positions of the whales can be updated by the following equations.

$$\vec{X}(t+1) = \vec{X}_{rand} - \vec{A}.\vec{D} \tag{8}$$

$$\vec{D} = |C.\vec{X}_{rand} - \vec{X}(t)| \tag{9}$$

where $\vec{X}_{rand}$ represents a random whale selected from the present population of whales.

In order to provide a proper trade-off between exploration and exploitation, the value of $\vec{A}$ is used. If the value is less than 1, exploitation is accomplished, else exploration.

$$\vec{X}(t+1) = \begin{cases} Equation\ 3, & |\vec{A}| < 1 \\ Equation\ 8, & |\vec{A}| \geq 1 \end{cases} \tag{10}$$

The value of a random number $p$ in [0,1] decides the type of movement (shrinking encircling or spiral motion) followed by the whales in WOA. If the value of $p < 0.5$ then shrinking encircling else spiral motion is undertaken.

$$\vec{X}(t+1) = \begin{cases} Shrinking\ encircling\ (Equation\ 10), & p < 0.5 \\ Spiral\ Motion\ (Equation\ 2), & p \geq 0.5 \end{cases} \tag{11}$$

During the entire process, the fitness of each whale is calculated using two prime objectives of FS – number of selected features and its classification accuracy. The ultimate goal of FS is to improve the classification accuracy and decrease the number of selected features. Hence, the fitness function is computed according to the following equation.

$$fitness_i = \alpha * acc_i + \beta * \frac{totFeat - |whale_i|}{totFeat} \tag{12}$$

where $fitness_i$, $acc_i$ define the fitness and classification accuracy of the $i^{th}$ whale respectively and $totFeat$ is total number of features present in the dataset.

There are multiple random parameters in WOA. For example, $\vec{A}$ and $\vec{C}$ in shrinking encircle mechanism, $\vec{l}$ for spiral shaped motion, $p$ which decides the selection of the searching procedure (shrinking encircling or spiral motion). Among these random variables, $p$ is the most important parameter because it guides the movement of the whales. If every time a random value for $p$ is selected, there may be some unwanted biasness in the number of times $p$ is less than or greater than 0.5. Hence, a systematic change in the value of $p$ is better suited as both types of movements are followed by the whales (other parameters do not necessarily need regular changes in their values). Inspired from [51], a chaotic approach has been introduced to change the value of $p$ in WOA. The value of $p$ over the iterations is generated using chaos functions. The nature of chaotic maps is unpredictable and random, but they also contain some element of regularity [73] which is needed to efficiently distribute the value of $p$ over [0,1]. Thus, chaos helps in bringing ergodicity in the deterministic dynamic system of WOA. In the present scenario, the choice of whale movement is guided by 4 chaotic maps – circular, logistics, piecewise and tent as given in Table 1. The balance between shrinking encircling and spiral shaped movements over the iterations is maintained by the chaos functions.

**Table 1:** Chaotic maps used in ECWSA.

| Sl. no. | Map Name | Map Equation |
|---|---|---|
| 1 | Circular | $p_{i+1} = (p_i + b - \left(\frac{a}{2*\pi}\right) * sin(2 * \pi * p_i))\%1$ |
| 2 | Logistics | $p_{i+1} = a * p_i * (1 - p_i)$ |
| 3 | Piecewise | $p_{i+1} = \begin{cases} p_i/a, & p_i < a \text{ and } p_i \geq 0 \\ (p_i - a)/(0.5 - a), & a \leq p_i \text{ and } p_i < 0.5 \\ (1 - a - p_i)/(0.5 - a), & p_i \geq 0.5 \text{ and } p_i < (1 - a) \\ (1 - p_i)/a, & p_i < 1 \text{ and } p_i \geq (1 - a) \end{cases}$ |
| 4 | Tent | $p_{i+1} = \begin{cases} p_i/0.7, & p_i < 0.7 \\ (10 * (1 - p_i))/3, & p_i < a \text{ and } p_i \geq 0 \end{cases}$ |

The different chaotic maps used in ECWSA is graphically presented in Figure 1 **(a-d):** Visualization of different chaotic maps used in the proposed method. . The visualization clearly shows the random nature of the mappings. The initial point of a mapping may have significant effect on its fluctuation patterns. The initial value for a chaotic map can be any value in the range [0,1] (or in [-1,1] depending on the range of the mapping). Our model has been tested by setting the initial point as 0.1, 0.3, 0.5 and 0.7. Out of these four values, our model works the best for 0.3. Hence, 0.3 has been selected as our initial point for all the maps. From Figure 1 **(a-d):** Visualization of different chaotic maps used in the proposed method. , it can be observed that the selected chaotic maps are distinguishable in nature which helps to properly peruse the effects of embedding different chaotic maps in the proposed method.

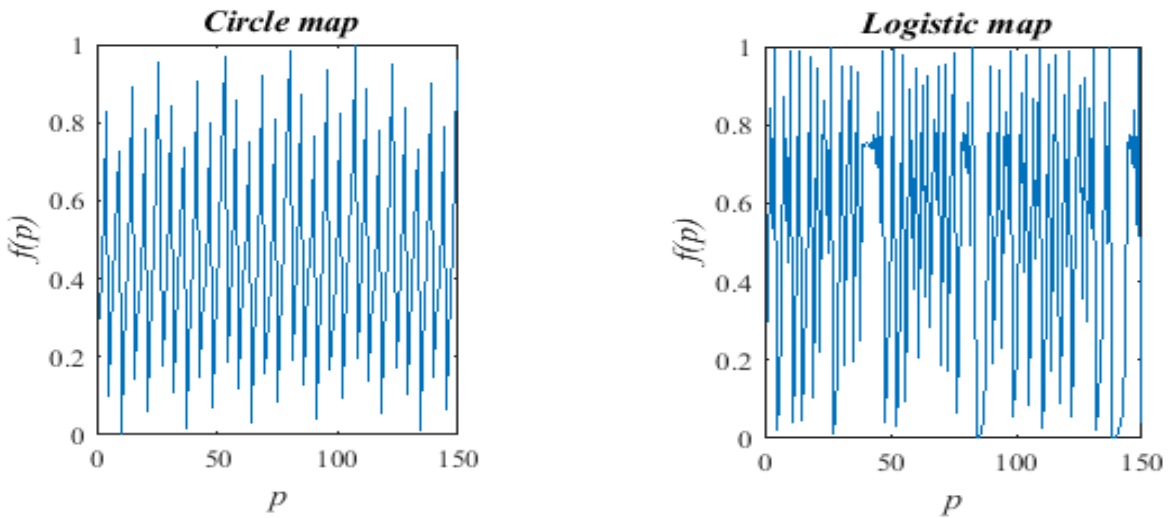

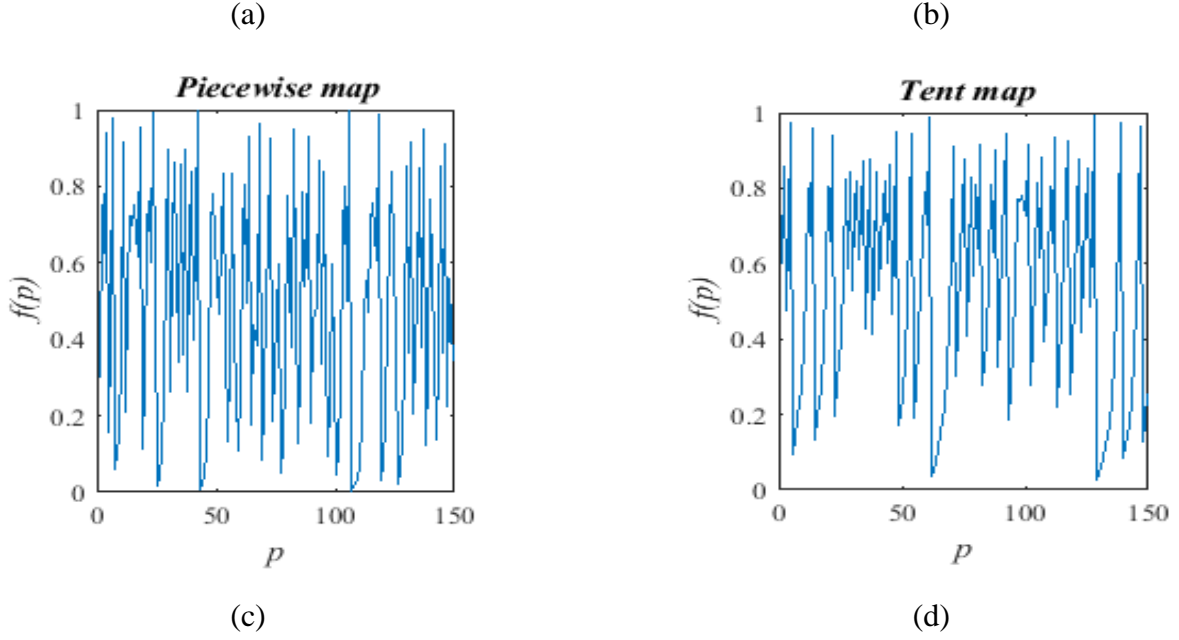

(a)            (b)

(c)            (d)

**Figure 1 (a-d):** Visualization of different chaotic maps used in the proposed method.

In real-life scenario, when the whales chase the prey using shrinking encircling or spiral motion, the prey keeps on moving. So, when the whales reach the position of the prey, the prey would have moved to some other place. Hence, the whales need to search in nearby locations to get to the exact location of the prey. To implement this scenario in the system of whales, taking inspiration from [74], a local search technique has also been applied to mimic the local searching behavior of the whales. This searching procedure uses mRMR to evaluate performance of the neighbors of the whales. This is a low-cost filter approach for enhancing local search capability. The diversity available in the population is utilized to generate the subset. Two random whales are selected and from that two new whales are generated using equations 13 and 14. The operators '−' and '∪' (used in equations 13 and 14) are set difference operator and union operator respectively.

$$\overrightarrow{X_i^1}(t+1) = \overrightarrow{X_i}(t) \cup (\overrightarrow{X_j}(t) - \overrightarrow{X_k}(t)) \qquad (13)$$

$$\overrightarrow{X_i^2}(t+1) = \overrightarrow{X_i}(t) - (\overrightarrow{X_j}(t) - \overrightarrow{X_k}(t)) \qquad (14)$$

The two new whales are compared with the initial whale ($X_i(t)$) and if any one of them has a higher fitness value (evaluated using mRMR called $mrmr\_fitness$) then the newly generated whale with the highest $mrmr\_fitness$ substitutes the original in the population. The mRMR based fitness of whale using mRMR is compared using equation 14. The pseudo code of this local search procedure in provided in Algorithm 1.

$$mrmr\_fitness = \left(\frac{1}{|X_i^a|}\sum_{x_i \in X_i^a} MI(x_i, class)\right) - \left(\frac{1}{|X_i^a|^2}\sum_{x_i, x_j \in X_i^a} MI(x_i, x_j)\right) \qquad (15)$$

**Procedure**: $loc\_search(\textbf{population of whales}, \textbf{mrmr\_fitness})$

1. $for\ i = 1\ to\ n\ do$
2.     $[r_1, r_2] = select\ two\ random\ whales\ from\ the\ population\ apart\ from\ whale_i$
3.     $Find\ the\ set\ difference\ between\ r_1\ and\ r_2\ as: dif_i = set\_difference(r_1, r_2)$
4.     $Form\ first\ neighbor\ as: neighbor_1 = whale_i \cup dif_i$
5.     $Form\ second\ neighbor\ as: neighbor_2 = whale_i - dif_i$
    // Calculate fitness of both the neighbors
6.     $fit(1) = mrmr\_fitness(neighbor_1)$
7.     $fit(2) = mrmr\_fitness(neighbor_2)$
    // Replace the current whale with the fittest neighbor if its fitness exceeds that of the current whale
8.     $if\ fit(1) > fitness(whale_i)$
9.         $whale_i = neighbor_1$
10.     $end\ if$
11.     $if\ fit(2) > fitness(whale_i)$
12.         $whale_i = neighbor_2$
13.     $end\ if$
14. $end\ for$

**Algorithm 1:** Pseudo code of the proposed local searching procedure using mRMR based fitness calculation (mentioned in equation 15).

So, the large number of diversification operations provide us with a good exploration and exploitation capacity but convergence is not enhanced by any of the operations. The diverging effects of the operations may lead to poor (non-converging) results. To counteract this and bring some balance between diversification and convergence, the size of the population is linearly decreased to drop the malnourished whales and to allow the fitter whales to pass to the next iteration. This model mimics the natural phenomenon of death of undernourished whales. At each iteration the population size is updated using equation 16. The value of $death$ is in the range of $[0, 1]$. It signifies the portion of the population which dies. In this way, the population size of the whales decreases over the iterations till a minimum size is reached which is known as the $base$. It restricts the number of whales to a predefined minimum value which is required for the search process to continue efficiently. In an intuitive way, it can be said that as there are very small number of whales at the end, all of them are surviving as they can get adequate food due to less competition. From searching perspective, at the later stages of iterations, it is considered that ECWSA algorithm has already found some of the near-optimal solutions in the search space as the survived whales and all of them can lead to better optima. So, they are not discarded from the solution space.

$$population\_size = \max(base, population\_size * (1 - death)) \qquad (16)$$

The entire workflow of the proposed method is represented in Figure 2. For better understanding, the pseudo code of the entire procedure is presented in Algorithm 2.

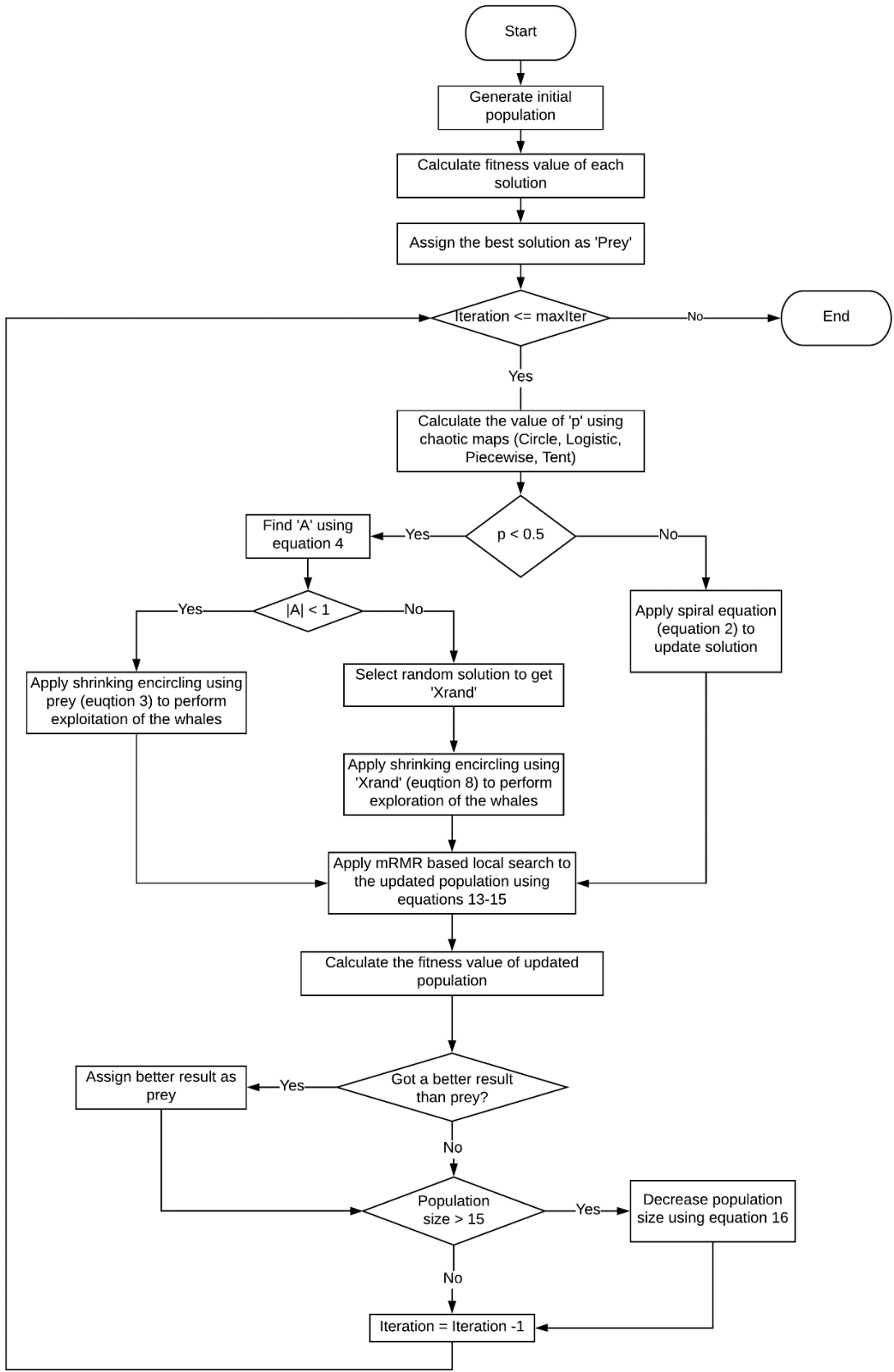

**Figure 2:** Flowchart of proposed feature selection algorithm called ECWSA.

**Algorithm:** Embedded Chaotic Whale Survival Algorithm

**Inputs**

Number of initial whales: $n$
Maximum number of iterations: $max\_iter$
Decrement ratio: $r$
Chaotic function: $chaos$
Movement selection parameter: $p$

**Output**

Prey or the fittest whale in the population

**Pseudo Code**

1. *Create a population of n initial whales*
2. $t \leftarrow 0$
3. $while(\mathbf{t} < \mathbf{max\_iter})\ do$
4.     *for $i = 1$ to $n$ do*
5.         *Calculate fitness of $whale_i$ as: $fit(i) = fitness(whale_i)$*
6.     *End for*
7.     *Sort the whales according to decreasing fitness values*
8.     *Select the fittest whale as the prey*
9.     *for $i = 1$ to $n$ do*
10.         *$p = chaos(p)$*
11.         *if$(\mathbf{p < 0.5})\ do$*
12.             *Calculate $\vec{A}$ accroding to Equation 4*
13.             *if$(|\vec{A}| < \mathbf{1})\ do$*

              // Exploitation using shrinking encircling mechanism

14.                 *Update position of $whale_i$ as directed in Equation 3*
15.             *else*

              // Exploration using shrinking encircling mechanism

16.                 *Select a random whale from the population*
17.                 *Update position of $whale_i$ as directed in Equation 8*
18.             *end if*
19.         *else*

            // Spiral Motion

20.             *Update position of $whale_i$ as directed in Equation 2*
21.         *end if*
22.     *end for*
23.     *Peform local search on the population of whales*
24.     *$n = \max(15, floor(r \times n))$*
25. *end while*

**Algorithm 2:** Pseudo code of ECWSA.

## 3.2 Time Complexity Analysis

To calculate the time complexity of the proposed ECWSA model, at first, the inputs to the algorithm need to be considered. The input parameters of the proposed model are:
**Initial population size:** $m$
**No. of iterations:** $n$
**Survival ratio:** $r$
**Number of features:** $num$

Let $k$ be the fixed population size referred to as $base$ in equation 15 to which the population of whales converges after some number of iterations according to the algorithm.
Now let us see the change in population size of the proposed model over the iteration. This description is provided in Table 2.

**Table 2:** Iteration-wise population variation in terms of the input parameters.

| Iteration number | Population size |
|---|---|
| 1 | $m$ |
| 2 | $m * r$ |
| 3 | $m * r^2$ |
| . | . |
| . | . |
| $i$ | $m * r^{i-1} = k$ |
| $i+1$ | $k$ |
| $i+2$ | $k$ |
| . | . |
| . | . |
| $n$ | $k$ |

From Table 2, a new equation (equation 17) is derived,
$$m * r^{i-1} = k \qquad (17)$$
After solving for $i$ in equation 17, the value of $i$ is $\log_r^{r*k/m}$

Hence, total number of evaluations
$$= m * (1 + r + r^2 + \ldots + r^{i-1}) + k * (n - i)$$
$$= m * \left(\frac{1 - r^i}{1 - r}\right) + k * (n - i)$$
$$= m * \left(\frac{1 - \frac{r*k}{m}}{1 - r}\right) + k(n - i)$$
$$= \frac{m - (r*k)}{1 - r} + k(n - i) \leq m * n$$

This is the final expression for the number of evaluations. The worst-case scenario occurs when $r = 1$ and $k = m$. Hence, if the time complexity of classification by the classifier is considered to be $O(num)$, the worst-case time complexity of the overall approach becomes $O(m * n * num)$. But the worst-case is very rare. So, most of the times, the time complexity is lesser than that.

## 4. Experimental Results

This section contains the results of the experiments conducted to evaluate the FS ability of the ECWSA. In section 4.1, the datasets used for experimentation are described followed by search for optimal values of parameters in section 4.2. Comparison with state-of-the-art methods is done in section 4.3 and in section 4.4 stability evaluation of the proposed method is described.

### 4.1 Dataset Description

The proposed FS algorithm has been tested on 18 popular UCI datasets. Details of the datasets are provided in Table 3. Depending on the number of classes, the datasets are divided into two categories-

- Two-class datasets
- Multi-class datasets

Out of 18 datasets, 13 are two-class datasets and rest 5 are multi-class datasets. 4 chaotic maps have been used to guide the movement of the whales. Depending on the serial number (say n) of the chaos function presented in Table 1, the corresponding ECWSA approach is named as ECWS-n. As 4 chaotic maps are used, the respective ECWSA versions are named as ECWSA-1, ECWSA-2, ECWSA-3 and ECWSA-4.

**Table 3:** Description of 18 UCI datasets used in evaluation of proposed method.

| Type of Dataset | Dataset | Number of attributes | Number of instances | Number of classes |
|---|---|---|---|---|
| Two-class | Breastcancer | 9 | 699 | 2 |
| | BreastEW | 30 | 569 | |
| | CongressEW | 16 | 435 | |
| | Exactly | 13 | 1000 | |
| | Exactly2 | 13 | 1000 | |
| | HeartEW | 13 | 270 | |
| | IonosphereEW | 34 | 351 | |
| | KrvskpEW | 36 | 3196 | |
| | M-of-n | 13 | 1000 | |
| | SonarEW | 60 | 208 | |
| | SpectEW | 22 | 267 | |
| | Tic-tac-toe | 9 | 958 | |
| | Vote | 16 | 300 | |
| Multi-class | WaveformEW | 40 | 5000 | 3 |
| | WineEW | 13 | 178 | 3 |
| | Lymphography | 18 | 148 | 4 |
| | PenglungEW | 325 | 73 | 7 |
| | Zoo | 16 | 101 | 7 |

### 4.2 Optimal Parameters

There are mainly two parameters present in the proposed approach – initial number of whales and number of generations. To find the optimal parameter values, IonosphereEW from two-class datasets and PenglungEW from multi-class datasets have been selected for experimentation. IonosphereEW is a well-known dataset used in the research community for binary classification. It consists of 34 attributes, 351 instances and classifies radar returns into "good" and "bad" classes depending on the presence of some kind of structure in the ionosphere which is quite interesting. On the other hand, PenglungEW contains 325 attributes and 73 instances. Due to this large number of features, the application of FS is more profound over PenglungEW. Based on said facts, IonosphereEW and PenglungEW are selected as the representatives of two-class and multi-class datasets respectively. Number of whales has been varied as 20, 40, 60, 80, 100 and number of generations has been changed as 10, 15, 20, 25, and 30. The K value of KNN classifier has been always kept fixed at 5 throughout the whole experimentation. Thus, a total of 25 parameter combination have been tested on each of 4 different versions of ECWSA. From the analysis of the results obtained for parameter variation, it can be observed that for 80 whales and 25 generations, the proposed model produces the best classification accuracy among all the 25 possible parameter combinations. The experimental outcomes for varying initial number of whales and number of generations are provided in Table 4. Hence, hereafter initial number of whales is set as 80 and number of generations as 25 for rest of the experimentations. Results of 20 runs are generated and the statistics are provided in Table 5 for all the datasets.

**Table 4:** Classification accuracy and percentage of features selected by different versions of ECWSA over PenglungEW and Ionosphere datasets for varying initial number of whales and number of generations.

| Initial number of whales | Number of generations | ECWSA version | PenglungEW Classification accuracy (in %) | Percentage of selected features | Ionosphere Classification accuracy (in %) | Percentage of selected features |
|---|---|---|---|---|---|---|
| 20 | 10 | 1 | 77.76 | 8.31 | 80.99 | 26.47 |
|  |  | 2 | 81.93 | 34.46 | 82.01 | 61.76 |
|  |  | 3 | 86.69 | 50.46 | 84.02 | 5.88 |
|  |  | 4 | 81.93 | 41.54 | 85.21 | 5.88 |
|  | 15 | 1 | 82.23 | 27.38 | 85.01 | 5.88 |
|  |  | 2 | 84.31 | 51.69 | 83.01 | 8.82 |
|  |  | 3 | 82.23 | 17.23 | 84.02 | 2.94 |
|  |  | 4 | 84.31 | 10.15 | 84.01 | 26.47 |
|  | 20 | 1 | 86.39 | 8.62 | 82 | 11.76 |
|  |  | 2 | 84.61 | 17.54 | 80.54 | 55.88 |
|  |  | 3 | 86.69 | 29.23 | 85.01 | 41.18 |
|  |  | 4 | 84.31 | 10.46 | 83.03 | 26.47 |
|  | 25 | 1 | 81.93 | 49.85 | 82.51 | 5.88 |
|  |  | 2 | 82.52 | 24.62 | 83.5 | 11.76 |
|  |  | 3 | 84.61 | 14.46 | 83.01 | 55.88 |
|  |  | 4 | 86.39 | 3.08 | 85.02 | 26.47 |
|  | 30 | 1 | 86.69 | 21.23 | 84 | 17.65 |
|  |  | 2 | 82.23 | 13.85 | 82.98 | 20.59 |
|  |  | 3 | 86.39 | 9.23 | 85.02 | 8.82 |
|  |  | 4 | 86.39 | 9.54 | 80.52 | 14.71 |
| 40 | 10 | 1 | 81.93 | 45.85 | 84.53 | 58.82 |
|  |  | 2 | 81.93 | 9.23 | 82.51 | 11.76 |
|  |  | 3 | 84.31 | 15.08 | 84.03 | 70.59 |
|  |  | 4 | 84.31 | 15.69 | 85.52 | 52.94 |
|  | 15 | 1 | 84.31 | 48 | 85 | 29.41 |
|  |  | 2 | 81.93 | 59.38 | 83.5 | 50 |
|  |  | 3 | 86.69 | 56 | 85.49 | 17.65 |
|  |  | 4 | 84.61 | 18.15 | 82.53 | 50 |
|  | 20 | 1 | 84.61 | 27.38 | 81.02 | 11.76 |
|  |  | 2 | 84.31 | 43.08 | 85.01 | 29.41 |
|  |  | 3 | 88.48 | 35.69 | 83.01 | 23.53 |
|  |  | 4 | 82.52 | 19.38 | 82.53 | 52.94 |
|  | 25 | 1 | 86.69 | 15.08 | 85.53 | 17.65 |
|  |  | 2 | 84.61 | 18.46 | 82.52 | 14.71 |
|  |  | 3 | 84.31 | 34.46 | 83.01 | 70.59 |
|  |  | 4 | 86.69 | 21.54 | 80.51 | 8.82 |
|  | 30 | 1 | 84.01 | 22.15 | 85.53 | 32.35 |
|  |  | 2 | 88.48 | 36.92 | 84.04 | 76.47 |
|  |  | 3 | 84.9 | 2.77 | 85.02 | 8.82 |

| Initial number of whales | Number of generations | ECWSA version | PenglungEW | | Ionosphere | |
|---|---|---|---|---|---|---|
| | | | Classification accuracy (in %) | Percentage of selected features | Classification accuracy (in %) | Percentage of selected features |
| 60 | 10 | 1 | 84.61 | 8.62 | 84.51 | 8.82 |
| | | 2 | 84.31 | 22.77 | 82.52 | 26.47 |
| | | 3 | 81.93 | 53.23 | 81.53 | 11.76 |
| | | 4 | 82.23 | 20.62 | 83.99 | 23.53 |
| | 15 | 1 | 86.39 | 11.69 | 82.01 | 52.94 |
| | | 2 | 84.61 | 29.54 | 85.53 | 61.76 |
| | | 3 | 88.77 | 18.15 | 84.02 | 85.29 |
| | | 4 | 81.93 | 12 | 83.54 | 47.06 |
| | 20 | 1 | 84.61 | 33.23 | 86.04 | 17.65 |
| | | 2 | 86.39 | 35.08 | 83.04 | 11.76 |
| | | 3 | 84.31 | 19.69 | 85.51 | 23.53 |
| | | 4 | 84.31 | 5.85 | 83.5 | 26.47 |
| | 25 | 1 | 84.61 | 32.62 | 85.04 | 8.82 |
| | | 2 | 84.61 | 22.15 | 87.51 | 17.65 |
| | | 3 | 84.61 | 30.15 | 83.53 | 52.94 |
| | | 4 | 86.99 | 52.92 | 81.5 | 91.18 |
| | 30 | 1 | 84.31 | 22.15 | 83.51 | 26.47 |
| | | 2 | 84.61 | 27.38 | 84.52 | 47.06 |
| | | 3 | 84.31 | 21.85 | 83.01 | 20.59 |
| | | 4 | 84.31 | 31.69 | 85.03 | 14.71 |
| 80 | 10 | 1 | 86.99 | 26.46 | 83.01 | 50 |
| | | 2 | 84.31 | 13.23 | 87.51 | 11.76 |
| | | 3 | 86.69 | 16.31 | 81.02 | 32.35 |
| | | 4 | 86.99 | 28 | 85.53 | 8.82 |
| | 15 | 1 | 84.31 | 29.54 | 82.98 | 17.65 |
| | | 2 | 84.31 | 59.69 | 83.53 | 61.76 |
| | | 3 | 84.61 | 32.31 | 86.99 | 5.88 |
| | | 4 | 84.31 | 29.85 | 82.53 | 23.53 |
| | 20 | 1 | 84.31 | 20.31 | 83.53 | 58.82 |
| | | 2 | 86.69 | 30.46 | 86.52 | 14.71 |
| | | 3 | 86.39 | 22.46 | 83 | 23.53 |
| | | 4 | 84.01 | 73.54 | 84.02 | 35.29 |
| | 25 | 1 | 86.69 | 34.46 | 85.49 | 20.59 |
| | | 2 | 88.39 | 25.54 | 87.21 | 23.53 |
| | | 3 | 88.99 | 8.62 | 87.51 | 38.24 |
| | | 4 | 87.69 | 38.15 | 87.21 | 20.59 |
| | 30 | 1 | 88.01 | 26.77 | 86.85 | 52.94 |
| | | 2 | 84.61 | 16 | 85.53 | 67.65 |
| | | 3 | 86.69 | 21.85 | 86.51 | 41.18 |
| | | 4 | 86.99 | 26.15 | 85.02 | 17.65 |
| 100 | 10 | 1 | 88.77 | 5.23 | 87.51 | 23.53 |
| | | 2 | 70.92 | 30.77 | 87.02 | 47.06 |
| | | 3 | 84.01 | 8.92 | 84.02 | 17.65 |
| | | 4 | 77.76 | 15.08 | 83.01 | 55.88 |
| | 15 | 1 | 84.31 | 12.31 | 84.52 | 26.47 |
| | | 2 | 73 | 36 | 82 | 41.18 |
| | | 3 | 79.55 | 21.54 | 83.01 | 44.12 |
| | | 4 | 78.06 | 24.92 | 85.02 | 50 |
| | 20 | 1 | 82.23 | 14.15 | 84.04 | 26.47 |
| | | 2 | 73 | 9.54 | 83.01 | 20.59 |
| | | 3 | 82.23 | 46.46 | 82.01 | 35.29 |
| | | 4 | 79.85 | 7.69 | 85.51 | 23.53 |
| | 25 | 1 | 79.85 | 13.85 | 86.99 | 23.53 |
| | | 2 | 75.68 | 21.23 | 85.53 | 23.53 |
| | | 3 | 79.55 | 24 | 85.53 | 32.35 |
| | | 4 | 82.23 | 28.92 | 85.02 | 20.59 |
| | 30 | 1 | 73.3 | 16.92 | 83.5 | 73.53 |
| | | 2 | 75.68 | 15.69 | 85.01 | 29.41 |
| | | 3 | 75.38 | 4.92 | 86 | 20.59 |
| | | 4 | 75.08 | 10.46 | 85.53 | 2.94 |
| | | | 81.63 | 15.08 | 83.02 | 38.24 |

### 4.3 Comparison with state-of-the-art

| ECWSA-1 | | | | ECWSA-2 | | | | ECWSA-3 | | | | |
|---|---|---|---|---|---|---|---|---|---|---|---|---|
| Minimum | Average | Standard Deviation | Maximum | Minimum | Average | Standard Deviation | Maximum | Minimum | Average | Standard Deviation | Maximum | Minim |
| 94.76 | 95.18 | 0.18 | 95.26 | 94.76 | 95.06 | 0.23 | 95.26 | 94.76 | 95.11 | 0.21 | 95.26 | 94. |
| 97.24 | 97.33 | 0.14 | 97.74 | 97.24 | **97.44** | 0.15 | 97.99 | 97.24 | 97.43 | 0.19 | 97.74 | 97. |
| 95.72 | 96.19 | 0.34 | 96.71 | 95.39 | **96.24** | 0.33 | 96.71 | 95.72 | 96.23 | 0.28 | 96.38 | 95. |
| 71.67 | 78.11 | 6.69 | 91 | 71.67 | **80.73** | 7.52 | 98.67 | 71.5 | 80.24 | 8.52 | 98.67 | 71. |
| 77 | **79.12** | 1.23 | 79.83 | 77 | 78.28 | 1.4 | 79.83 | 77 | 78.59 | 1.37 | 79.83 | 77 |
| 84.66 | 85.56 | 0.34 | 85.71 | 85.19 | 85.61 | 0.21 | 85.71 | 84.66 | 85.5 | 0.39 | 85.71 | 85. |
| 84.04 | 86.72 | 1.58 | 87.53 | 83.48 | 85.59 | 1.23 | 89.51 | 84.03 | **86.99** | 1.2 | 89.51 | 83. |
| 93.22 | 93.92 | 0.59 | 95.72 | 93.01 | 94.09 | 0.77 | 96.19 | 93.07 | **94.61** | 0.88 | 94.37 | 92. |
| 85.78 | **87.39** | 0.73 | 89.65 | 85.67 | 87.24 | 1.04 | 89.65 | 85.73 | 87.3 | 1.16 | 90.6 | 84. |
| 89.67 | 92.13 | 1.95 | 97 | 90 | 93.11 | 2.37 | 97 | 90.33 | **93.84** | 2.71 | 97 | 89. |
| 84.23 | 87.66 | 1.71 | 93.15 | 84.23 | 88.02 | 1.87 | 91.37 | 84.52 | **88.65** | 1.9 | 91.07 | 84. |
| 74.47 | 76.38 | 1.33 | 79.43 | 73.05 | 76.67 | 1.71 | 80.14 | 74.47 | **76.88** | 1.59 | 80.85 | 74. |
| 77.66 | 79.88 | 1.35 | 81.59 | 76.37 | 79.68 | 1.4 | 81.59 | 76.56 | **79.9** | 1.49 | 81.59 | 76. |
| 78.78 | **78.78** | 0 | 78.78 | 78.09 | 78.75 | 0.15 | 78.78 | 78.09 | 78.75 | 0.15 | 78.78 | 78. |
| 94.44 | 94.97 | 0.21 | 95.56 | 94.44 | 95.03 | 0.28 | 95.56 | 95 | 95.06 | 0.17 | 96.11 | 95 |
| 78.97 | 79.85 | 0.53 | 81.74 | 79 | 80.14 | 0.76 | 81.49 | 78.8 | **80.22** | 0.71 | 81.43 | 78. |
| 97.74 | 98.02 | 0.35 | 98.52 | 97.74 | **98.31** | 0.32 | 98.52 | 97.74 | 98.13 | 0.37 | 98.52 | 97. |
| 96.82 | 98.7 | 0.83 | 100 | 98.15 | **99.35** | 0.81 | 100 | 98.15 | 99.27 | 0.81 | 100 | 96. |

In order to establish the superiority of the proposed model, the results obtained by ECWSA have been compared with some state-of-the-art FS methods. For proper evaluation, a fixed environment was used for experimentation. During the entire experimentation, KNN classifier is used for classification and the $K$ value is set to 5. For evaluation of candidate solutions, each dataset has been divided into $K$-fold cross validation where $K-1$ folds have been used for training and validation while the remaining fold has been used for testing. The results of other popular metaheuristic algorithms used in the comparison are taken from [23]. The comparison results show that the proposed ECWSA outperforms both recently developed versions of WOA as well as some state-of-the-art techniques used for FS like ALO, GA, PSO and Wrapper-Filter Ant Colony Optimization based Feature Selection (WFACOFS – a wrapper filter version of ACO). The other versions of WOA used for comparison are WOA with crossover and mutation (WOA-CM), WOA using tournament selection (WOA-T) and WOA using roulette selection (WOA-R).

**Table 6** and Table 7 show the comparison results in terms of classification accuracy and percentage of features selected respectively. In 11 datasets, the proposed algorithm outperforms other contemporaries in terms of average classification accuracy. WOA-CM appears to be the second-best performer among these methods achieving highest accuracies for 5 remaining datasets. WFACOFS is the third best when solving two datasets. The comparison of classification accuracy clearly shows the applicability of the proposed model.

From Table 7, it can be seen that ECWSA is able to reduce the feature dimension to a significant extent and the reduction is better than its contemporaries in 13 cases. Thus, the proposed model is able to increase the classification accuracy and reduce the feature dimension at the same time which are the two criteria of FS. It proves the effectiveness of ECWSA as a FS model. However, ECWSA does not achieve a much higher accuracy as compared to other algorithms. This is due to two shortcomings. For one, the dimensionally reduction ability of the algorithm is much more than accuracy increment. For another, the computational complexity of this algorithm is high due to the use of local search.

The ECWSA, as seen from Table 5, has a low standard deviation (less than 1.5 for most datasets). The use of the death to improve the convergence of the solutions helps in achieving this. Moreover, our enhanced exploitation and exploration abilities help to achieve good accuracy in most datasets, 11 out of 18 datasets. The fact that this accuracy is achieved using a lower number of average features (for 13 datasets out of 18) shows that the search space is better explored. Computation in our algorithm is not increased substantially due to the use of the local search since a filter method is used to determine the fitness. Therefore, our algorithm without a significant rise in computation has a better balance between exploration, exploitation and convergence.

**Table 5:** Minimum, maximum, average and standard deviation of the classification accuracies obtained by the four variants of proposed ECWSA over different datasets.

| Datasets | ECWSA-1 | | | | ECWSA-2 | | | | ECWSA-3 | | | | ECWSA-4 | | | |
|---|---|---|---|---|---|---|---|---|---|---|---|---|---|---|---|---|
| | Maximum | Minimum | Average | Standard Deviation | Maximum | Minimum | Average | Standard Deviation | Maximum | Minimum | Average | Standard Deviation | Maximum | Minimum | Average | Standard Deviation |
| Breastcancer | 95.26 | 94.76 | 95.18 | 0.18 | 95.26 | 94.76 | 95.06 | 0.23 | 95.26 | 94.76 | 95.11 | 0.21 | 95.26 | 94.76 | **95.21** | 0.13 |
| BreastEW | 97.74 | 97.24 | 97.33 | 0.14 | 97.74 | 97.24 | **97.44** | 0.15 | 97.99 | 97.24 | 97.43 | 0.19 | 97.74 | 97.24 | 97.38 | 0.15 |
| CongressEW | 96.71 | 95.72 | 96.19 | 0.34 | 96.71 | 95.39 | **96.24** | 0.33 | 96.71 | 95.72 | 96.23 | 0.28 | 96.38 | 95.72 | 96.23 | 0.24 |
| Exactly | 91 | 71.67 | 78.11 | 6.69 | 91 | 71.67 | **80.73** | 7.52 | 98.67 | 71.5 | 80.24 | 8.52 | 98.67 | 71.17 | 78.09 | 8.57 |
| Exactly2 | 79.83 | 77 | **79.12** | 1.23 | 79.83 | 77 | 78.28 | 1.4 | 79.83 | 77 | 78.59 | 1.37 | 79.83 | 77 | 78.9 | 1.27 |
| HeartEW | 85.71 | 84.66 | 85.56 | 0.34 | 85.71 | 85.19 | 85.61 | 0.21 | 85.71 | 84.66 | 85.5 | 0.39 | 85.71 | 85.19 | **85.63** | 0.19 |
| IonosphereEW | 89.51 | 84.04 | 86.72 | 1.58 | 87.53 | 83.48 | 85.59 | 1.23 | 89.51 | 84.03 | **86.99** | 1.2 | 89.51 | 83.04 | 86.79 | 1.48 |
| KrvskpEW | 95.52 | 93.22 | 93.92 | 0.59 | 95.72 | 93.01 | 94.09 | 0.77 | 96.19 | 93.07 | **94.61** | 0.88 | 94.37 | 92.39 | 93.53 | 0.57 |
| Lymphography | 88.64 | 85.78 | **87.39** | 0.73 | 89.65 | 85.67 | 87.24 | 1.04 | 89.65 | 85.73 | 87.3 | 1.16 | 90.6 | 84.85 | 87.02 | 1.45 |
| M-of-n | 97 | 89.67 | 92.13 | 1.95 | 97 | 90 | 93.11 | 2.37 | 97 | 90.33 | **93.84** | 2.71 | 97 | 89.33 | 92.47 | 2.83 |
| PenglungEW | 90.77 | 84.23 | 87.66 | 1.71 | 93.15 | 84.23 | 88.02 | 1.87 | 91.37 | 84.52 | **88.65** | 1.9 | 91.07 | 84.23 | 87.63 | 1.77 |
| SonarEW | 78.72 | 74.47 | 76.38 | 1.33 | 79.43 | 73.05 | 76.67 | 1.71 | 80.14 | 74.47 | **76.88** | 1.59 | 80.85 | 74.47 | 76.84 | 1.71 |
| SpectEW | 81.59 | 77.66 | 79.88 | 1.35 | 81.59 | 76.37 | 79.68 | 1.4 | 81.59 | 76.56 | **79.9** | 1.49 | 81.59 | 76.56 | 79.84 | 1.52 |
| Tic-tac-toe | 78.78 | 78.78 | **78.78** | 0 | 78.78 | 78.09 | 78.75 | 0.15 | 78.78 | 78.09 | 78.75 | 0.15 | 78.78 | 78.09 | 78.75 | 0.15 |
| Vote | 95.56 | 94.44 | 94.97 | 0.21 | 95.56 | 94.44 | 95.03 | 0.28 | 95.56 | 95 | 95.06 | 0.17 | 96.11 | 95 | **95.08** | 0.26 |
| WaveformEW | 81.17 | 78.97 | 79.85 | 0.53 | 81.74 | 79 | 80.14 | 0.76 | 81.49 | 78.8 | **80.22** | 0.71 | 81.43 | 78.91 | 80.18 | 0.65 |
| Wine | 98.52 | 97.74 | 98.02 | 0.35 | 98.52 | 97.74 | **98.31** | 0.32 | 98.52 | 97.74 | 98.13 | 0.37 | 98.52 | 97.74 | 98.02 | 0.35 |
| Zoo | 100 | 96.82 | 98.7 | 0.83 | 100 | 98.15 | **99.35** | 0.81 | 100 | 98.15 | 99.27 | 0.81 | 100 | 96.82 | 98.95 | 0.92 |

**Table 6:** Average classification accuracy of proposed FS methods for datasets in comparison to some state-of-the-art methods.

| Dataset | CLASSIFICATION ACCURACY (in %) | | | | | | | | | | | | |
|---|---|---|---|---|---|---|---|---|---|---|---|---|---|
| | WOA | WOA-T | WOA-R | WOA-CM | ALO | GA | PSO | HGAFS | WFACOFS | ECWSA-1 | ECWSA-2 | ECWSA-3 | ECWSA-4 |
| Breastcancer | 95.71 | 95.9 | 95.76 | 96.83 | 96.1 | 95.5 | 95.4 | 92.00 | **98.7** | 95.18 | 95.06 | 95.11 | 95.21 |
| BreastEW | 95.53 | 94.98 | 95.07 | 97.07 | 93 | 93.8 | 94.1 | 95.73 | 97.33 | 97.33 | **97.44** | 97.43 | 97.38 |
| CongressEW | 92.96 | 91.47 | 91.06 | 95.6 | 92.9 | 93.8 | 93.7 | 92.44 | 96 | 96.19 | **96.24** | 96.23 | 96.23 |
| Exactly | 75.76 | 73.96 | 76.33 | **100** | 66 | 66.6 | 68.4 | 69.83 | 75 | 78.11 | 80.73 | 80.24 | 78.09 |
| Exactly2 | 69.85 | 69.94 | 69.07 | 74.21 | 74.5 | 75.7 | 74.6 | 74.00 | 74 | **79.12** | 78.28 | 78.59 | 78.9 |
| HeartEW | 76.33 | 76.52 | 76.33 | 80.67 | 82.6 | 82.2 | 78.4 | 78.31 | 85.56 | 85.56 | 85.61 | 85.5 | **85.63** |
| IonosphereEW | 89.01 | 88.44 | 88.01 | 92.56 | 86.6 | 83.4 | 84.3 | 75.49 | **95** | 86.72 | 85.59 | 86.99 | 86.79 |
| KrvskpEW | 91.51 | 89.65 | 90.18 | **97.18** | 95.6 | 92.3 | 94.2 | 79.82 | 94 | 93.92 | 94.09 | 94.61 | 93.53 |
| Lymphography | 78.58 | 77.86 | 75.95 | 85.18 | 78.7 | 70.8 | 69.3 | 77.15 | 80 | **87.39** | 87.24 | 87.3 | 87.02 |
| M-of-n | 85.4 | 83.89 | 86.03 | **99.14** | 86.4 | 92.7 | 86.4 | 88.50 | 91 | 92.13 | 93.11 | 93.84 | 92.47 |
| PenglungEW | 72.97 | 73.65 | 71.22 | 79.19 | 62.7 | 69.6 | 72 | 74.70 | 86.33 | 87.66 | 88.02 | **88.65** | 87.63 |
| SonarEW | 85.43 | 86.11 | 85.72 | **91.88** | 73.8 | 72.6 | 74 | 64.54 | 53.88 | 76.38 | 76.67 | 76.88 | 76.84 |
| SpectEW | 78.77 | 79.22 | 77.87 | **86.57** | 80.1 | 77.5 | 76.9 | 65.02 | 76.9 | 79.88 | 79.68 | 79.9 | 79.84 |
| Tic-tac-toe | 75.11 | 73.63 | 74.98 | 78.54 | 72.5 | 71.3 | 72.8 | 74.08 | 78.75 | **78.78** | 78.75 | 78.75 | 78.75 |
| Vote | 93.87 | 93.5 | 93.23 | 93.87 | 91.7 | 89.4 | 89.4 | 89.44 | 93 | 94.97 | 95.03 | 95.06 | **95.08** |
| WaveformEW | 71.27 | 71.01 | 71.21 | 75.33 | 77.3 | 76.7 | 76.1 | 80.74 | 74 | 79.85 | 80.14 | **80.22** | 80.18 |
| Wine | 92.81 | 92.81 | 92.58 | 95.9 | 91.1 | 93.3 | 95 | 93.13 | 97.6 | 98.02 | **98.31** | 98.13 | 98.02 |
| Zoo | 96.47 | 96.47 | 95.69 | 98.04 | 90.9 | 88.4 | 83.4 | 94.97 | 80 | 98.7 | **99.35** | 99.27 | 98.95 |

**Table 7:** Average percentage of feature selected by proposed FS methods for different datasets in comparison to some state-of-the-art methods.

| Dataset | PERCENTAGE OF SELECTED FEATURES | | | | | | | | | | | | |
|---|---|---|---|---|---|---|---|---|---|---|---|---|---|
| | WOA | WOA-T | WOA-R | WOA-CM | ALO | GA | PSO | HGAFS | WFACOFS | ECWSA-1 | ECWSA-2 | ECWSA-3 | ECWSA-4 |
| Breastcancer | 59.4 | 66.1 | 62.8 | 47.8 | 69.8 | 56.6 | 63.6 | 60.00 | 67.00 | 48.5 | **46.5** | 47 | 51.5 |
| BreastEW | 69.2 | 68.5 | 72.5 | 52.7 | 53.6 | 54.5 | 55.2 | 86.67 | 64.83 | 53.5 | **50** | 52.33 | 50.33 |
| CongressEW | 64.7 | 64.1 | 56.3 | 40.3 | 43.6 | 41.4 | 42.7 | 62.50 | 49.69 | 35 | 37.5 | 40 | **26.56** |
| Exactly | 83.1 | 82.7 | 75.8 | **46.5** | 50.9 | 83.2 | 75 | 92.31 | 69.23 | 55 | 48.85 | 51.54 | 53.46 |
| Exactly2 | 44.2 | 69.2 | **20.4** | 40.4 | 82.3 | 47.5 | 47.5 | 61.54 | 50.62 | 69.23 | 59.23 | 64.62 | 70.38 |
| HeartEW | 66.5 | 64.6 | 58.8 | **53.5** | 79.3 | 73 | 61.1 | 84.62 | 67.31 | 72.31 | 73.46 | 75 | 69.23 |
| IonosphereEW | 63.1 | 59.4 | 55 | 42.4 | **27.7** | 50.9 | 56.4 | 58.82 | 28.09 | 27.79 | 31.18 | 29.26 | 28.68 |
| KrvskpEW | 77.5 | 74.2 | 76.8 | 51.5 | 68.6 | 62.3 | 57.8 | 86.11 | 65.50 | **36.25** | 42.5 | 49.03 | 44.58 |
| Lymphography | 58.6 | 51.9 | 54.4 | 45.6 | 61.4 | 61.4 | 49.9 | 88.89 | 67.39 | **42.78** | 53.33 | 46.39 | 54.44 |
| M-of-n | 75.4 | 81.2 | 79.6 | 46.2 | 85.2 | 52.5 | 69.5 | 92.31 | 78.69 | **38.46** | 38.85 | 53.46 | 40 |
| PenglungEW | 44.4 | 47.2 | 36 | 39.4 | 50.5 | 54.5 | 55 | 58.46 | 63.78 | **19** | 33.52 | 25.62 | 28.54 |
| SonarEW | 72.3 | 63.7 | 66.8 | 59.4 | 63.2 | 55.5 | 52 | 83.33 | 60.50 | **33.42** | 37.25 | 34.92 | 38.25 |
| SpectEW | 55 | 52.4 | 35.9 | 36.6 | 73.4 | 53.4 | 56.8 | 77.27 | 48.64 | 35.45 | 35 | 30.68 | **30** |
| Tic-tac-toe | 73.9 | 76.1 | 79.4 | 76.7 | 77.7 | 76.1 | **73.4** | 88.89 | 86.67 | 86.11 | 85 | 88.89 | 89.44 |
| Vote | 46.3 | 51.3 | 43.1 | 46.3 | 59.5 | 41.4 | 55 | 62.50 | 57.19 | 36.88 | 35 | **34.69** | 37.81 |
| WaveformEW | 83 | 84.3 | 85.6 | 63.5 | 89.3 | 63.2 | 56.8 | 60.00 | 59.38 | **35.38** | 41 | 38 | 38.75 |
| Wine | 68.1 | 68.5 | 71.9 | 52.3 | 82.3 | 66.4 | 64.3 | 76.92 | 57.54 | **48.46** | 52.69 | 49.23 | 55.77 |
| Zoo | 61.9 | 73.1 | 74.7 | 52.3 | 87.3 | 63.2 | 60.9 | 62.50 | 53.63 | 49.38 | 54.37 | 58.44 | **43.13** |

## 4.4 Stability Evaluation

In order to check the stability of the proposed approach in providing efficient FS, again IonosphereEW and PenglungEW datasets have been selected due to the reasons mentioned in section 4.2. Boxplots of the classification accuracies produced by all 4 versions of ECWSA over these two datasets are drawn. The boxplots are represented in **Error! Reference source not found.** and Figure 4 respectively which clearly confirm the stability of our proposed FS method. From the boxplot, it can be seen that the classification values of different candidate solutions are fairly distributed around the mean values. Convergence of ECWSA is verified by plotting the average accuracy in each iteration for runs. Figure 5 and Figure 6 are the convergence graph for datasets IonosphereEW and PenglungEW respectively. From both the figures, it can be seen that the classification ability of the candidate solutions gradually increased for all the four versions of ECWSA over the iterations. This visualization strengthens the proposed model's ability to alleviate the candidate solutions over different iterations.

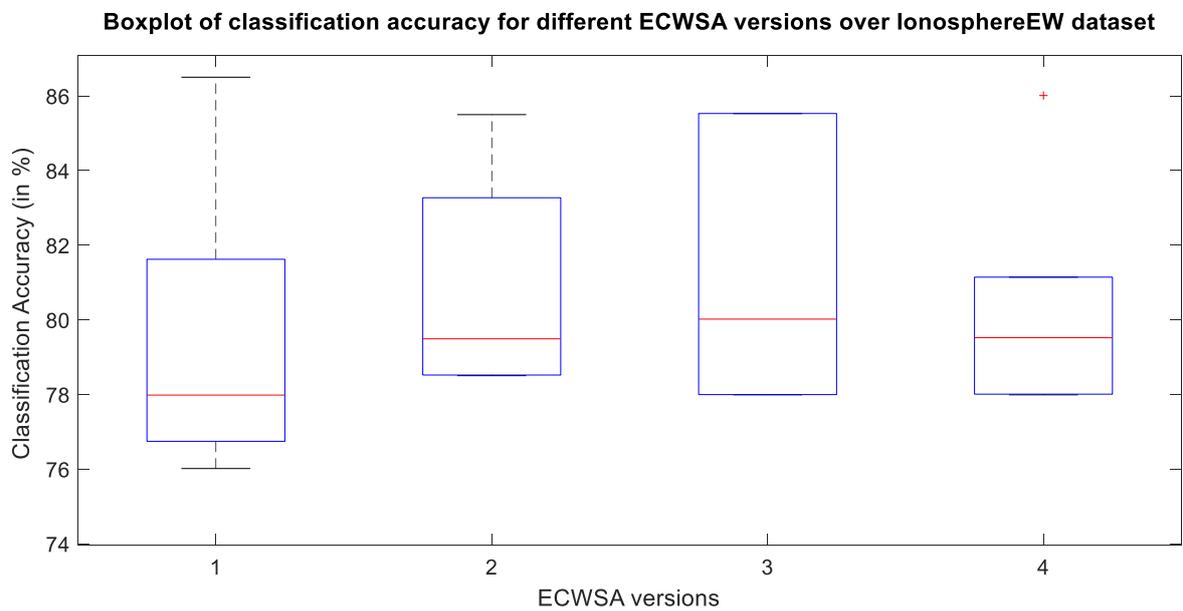

**Figure 3:** Box plot of classification accuracy obtained by different versions of ECWSA over IonosphereEW dataset.

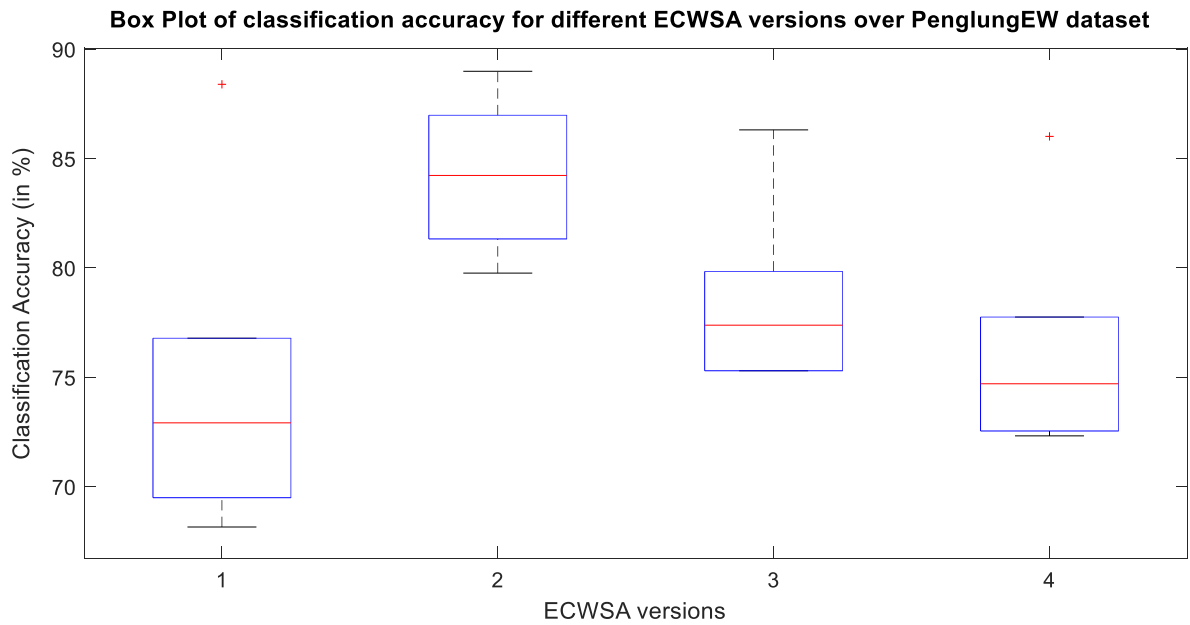

**Figure 4:** Box plot of classification accuracy obtained by different version of ECWSA over PenglungEW dataset.

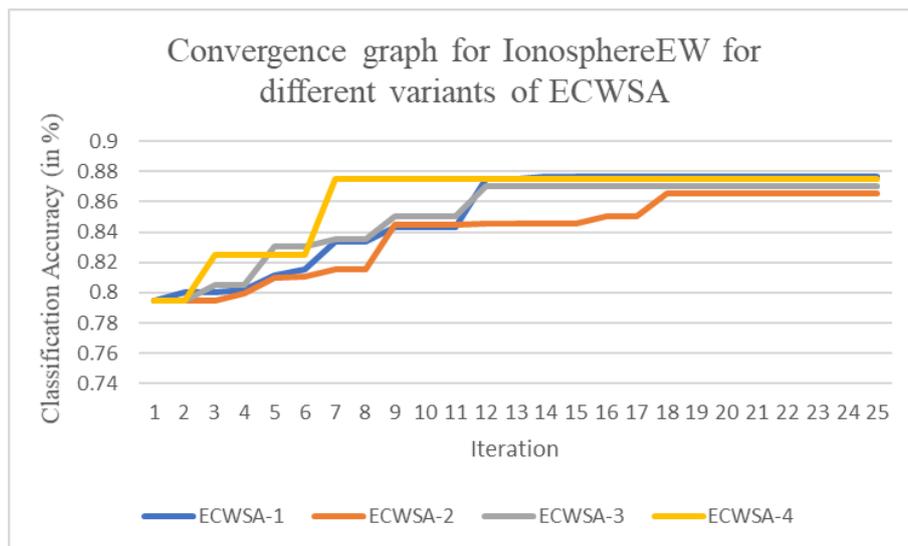

**Figure 5:** Convergence graph for IonosphereEW dataset. The classification accuracy of best in population plotted against the iteration number.

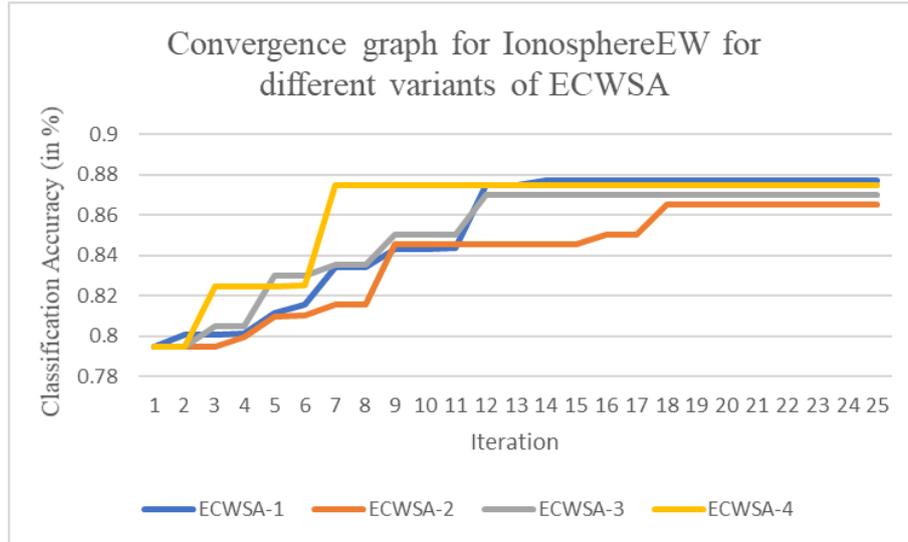

**Figure 6:** Convergence graph for PenglungEW dataset. The classification accuracy of best in population plotted against the iteration number.

From the experimental outcomes of, it is clearly visible that the proposed model namely ECWSA outperforms most of its contemporaries. There are several reasons that can be attributed to the success of ECWSA as an efficient FS model. First of all, mRMR based filter method allows the whales to explore a larger portion of the search space without requiring much time for the computation. It also helps to insert data inherent properties in the selection of features. Secondly, the introduction of chaos brings certain degree of randomness in the deterministic dynamic system of the whales. The chaotic mappings help the whales to choose the type of movement (shrinking encircling or spiral motion) where both the types have equal probability of selection. The death of weak whales at the end of each iteration enhances faster convergence thereby replicating the real phenomena as well as providing faster solutions. These novel approaches implemented in ECWSA combinedly place it ahead of its contemporaries in terms of FS ability.

### 4.5 Robustness Evaluation

In addition to the standard UCI datasets, ECWSA has been evaluated on various Microarray datasets [75] to check the robustness of the overall model in case of large dimensional feature sets. The description of the datasets used for the robustness testing is presented in Table 8. The corresponding results and comparison are tabulated in Tables 9 and 10.

**Table 8:** Descriptions of the 7 Microarray datasets used for testing robustness of ECWSA.

| Dataset | Number of features | Number of samples | Number of classes |
|---|---|---|---|
| AMLGSE2191 | 12616 | 54 | 2 |
| Colon | 7464 | 36 | 2 |
| DLBCL | 7070 | 77 | 2 |
| Leukaemia | 5147 | 72 | 2 |
| Prostate | 12533 | 102 | 2 |
| MLL | 12533 | 72 | 3 |
| SRBCT | 2308 | 83 | 4 |

**Table 9:** The results obtained for 4 different versions of ECWSA over Microarray datasets. The accuracies obtained without any FS (entire dataset) are also provided.

| Microarray Dataset | Accuracy on entire dataset (%) | Original feature dimension | ECWSA-1 | | ECWSA-2 | | ECWSA-3 | | ECWSA-4 | |
|---|---|---|---|---|---|---|---|---|---|---|
| | | | Accuracy (%) | Number of selected features | Accuracy (%) | Number of selected features | Accuracy (%) | Number of selected features | Accuracy (%) | Number of selected features |
| AMLGSE2191 | 51.85 | 12616 | 96.67 | 17 | 100 | 9 | 95.83 | 16 | 95.83 | 18 |
| Colon | 88.89 | 7464 | 100 | 36 | 100 | 41 | 100 | 30 | 100 | 43 |
| DLBCL | 76.92 | 7070 | 100 | 29 | 100 | 24 | 100 | 26 | 100 | 31 |
| Leukaemia | 83.78 | 5147 | 97.22 | 7 | 100 | 8 | 100 | 4 | 97.22 | 5 |
| Prostate | 62.75 | 12533 | 96.3 | 16 | 98.15 | 16 | 96.3 | 9 | 96.3 | 19 |
| MLL | 68.57 | 12533 | 100 | 16 | 100 | 17 | 100 | 8 | 100 | 15 |
| SRBCT | 85 | 2308 | 100 | 45 | 100 | 32 | 100 | 34 | 100 | 30 |

**Table 10:** Comparison of computed Microarray results with some state-of-the-art FS algorithms. The number features selected are provided in brackets at the side of the accuracy.

| Datasets | Accuracy (in %) | | | | | | |
|---|---|---|---|---|---|---|---|
| | GA | MA | WFACOFS | ECWSA-1 | ECWSA-2 | ECWSA-3 | ECWSA-4 |
| AMLGSE2191 | 100(98) | 100(91) | 96.3(17) | 96.67(17) | **100.00(9)** | 95.83(16) | 95.83(18) |
| Colon | 100(81) | 100(81) | **100(3)** | 100.00(36) | 100.00(41) | 100.00(30) | 100.00(43) |
| DLBCL | 100(88) | 100(105) | **100(3)** | 100.00(29) | 100.00(24) | 100.00(26) | 100.00(31) |
| Leukaemia | 100(85) | 100(65) | 100(5) | 97.22(7) | 100.00(8) | **100.00(4)** | 97.22(5) |
| Prostate | 100(99) | 100(107) | **100(22)** | 96.30(16) | 98.15(16) | 96.30(9) | 96.30(19) |
| MLL | 100(94) | 100(80) | 100(25) | 100.00(16) | 100.00(17) | **100.00(8)** | 100.00(15) |
| SRBCT | 100(78) | 100(50) | **100(19)** | 100.00(45) | 100.00(32) | 100.00(34) | 100.00(30) |

The four different variants of ECWSA have been applied to seven microarray datasets outlined in Table 8. The results obtained for all the variants are provided in Table 9. This table also contains the classification accuracy for the entire dataset prior to FS. It can be seen that all the variants of ECWSA have performed exceptionally good. In four out of seven datasets (Colon, DLBCL, MLL, SRBCT); all algorithms have been able to achieve 100% classification accuracy. In case of Leukaemia, two algorithms provides 100% accuracy, while for AMLGSE2191, one algorithm gives 100% accuracy. None of the variants is able to achieve 100% accuracy for Prostate but all of them obtained accuracies higher than 96%. If the focus is on number of features used to achieve these accuracies, in every case, they have used less than 2% of the total number of features in the datasets. Finally, the results obtained for ECWSA variants are compared with the results of some of the state-of-the-art FS algorithms namely GA [76], Memetic Algorithm (MA) [17], [77] and WFACOFS [20].

## 5. Conclusion

In this work, a new method for FS which is based on WOA has been proposed. The algorithm called ECWSA or Embedded Chaotic Whale Survival Algorithm is a filter-wrapper algorithm which uses a local search mechanism aided with mRMR as a performance evaluation tool. Better representation of whale foraging has been done by incorporating the death of less fit individuals. Moreover, chaos has been used to select which whales undergo shrinking encircling and which perform spiral motion. This technique helps to better avail the explorative capacity of WOA by incorporating pinpointed local search. This prevents exploration affecting local search and thereby leading to better balance. Boxplots is provided to display the stability of the proposed approach. The results for ECWSA show a significant improvement in terms of FS in 11 datasets out of 18. One of the shortcomings of this algorithm is the computational complexity required to perform the local search and chaos-based movements. In future, ECWSA could be hybridized with other population-based FS approaches like ACO, PSO etc. A filter-based classifier could be used too to perform selection, this would greatly reduce computational complexity. ECWSA has been applied on microarray data in this work. A deeper analysis of the selections done by ECWSA and their biological impact can be studied. The proposed algorithm can also be applied to some real-world problems like handwritten word or digit recognition, graphology applications, sleep deprivation detection and so on. Further analysis of impact of use of other classifiers like Neural Networks or Radom Forest can also be made.

## References

[1]   J. Han, J. Pei, and M. Kamber, *Data mining: concepts and techniques*. Elsevier, 2011.
[2]   R. Jensen, "Combining rough and fuzzy sets for feature selection." Citeseer, 2005.
[3]   I. Guyon and A. Elisseeff, "An introduction to variable and feature selection," *J. Mach. Learn. Res.*, vol. 3, no. Mar, pp. 1157–1182, 2003.
[4]   J. Huang, "A hybrid genetic algorithm for feature selection wrapper based on mutual information," vol. 28, pp. 1825–1844, 2007.
[5]   S. Malakar, M. Ghosh, S. Bhowmik, R. Sarkar, and M. Nasipuri, "A GA based hierarchical feature selection approach for handwritten word recognition," *Neural Comput. Appl.*, pp. 1–20, 2019.
[6]   H. Y. Markid, B. Z. Dadaneh, and M. E. Moghaddam, "Bidirectional ant colony optimization for feature selection," in *2015 The International Symposium on Artificial Intelligence and Signal Processing (AISP)*, 2015, pp. 53–58.
[7]   S. Kashef and H. Nezamabadi-pour, "An advanced ACO algorithm for feature subset selection," *Neurocomputing*, vol. 147, no. 1, pp. 271–279, 2015.
[8]   J. Wei *et al.*, "A BPSO-SVM algorithm based on memory renewal and enhanced mutation mechanisms for feature selection," *Appl. Soft Comput. J.*, vol. 58, pp. 176–192, 2017.
[9]   H. Liu and H. Motoda, *Computational methods of feature selection*, vol. 20071386. CRC Press, 2007.
[10]  P. Mitra, C. A. Murthy, and S. K. Pal, "Unsupervised feature selection using feature similarity," *IEEE Trans. Pattern Anal. Mach. Intell.*, vol. 24, no. 3, pp. 301–312, 2002.
[11]  W.-Q. Shang, Y.-L. Qu, H.-K. Huang, H.-B. Zhu, Y.-M. Lin, and H.-B. Dong, "Fuzzy knn text classifier based

on gini index," *J. Guang xi Norm. Univ. Nat. Sci. Ed.*, vol. 24, no. 4, pp. 87–90, 2006.

[12] J. Biesiada and W. Duch, "Feature Selection for High-Dimensional Data — A Pearson Redundancy Based Filter," 2007, pp. 242–249.

[13] N. Sánchez-Maroño, A. Alonso-Betanzos, and M. Tombilla-Sanromán, "Filter Methods for Feature Selection – A Comparative Study," in *Intelligent Data Engineering and Automated Learning - IDEAL 2007*, Berlin, Heidelberg: Springer Berlin Heidelberg, pp. 178–187.

[14] R. Guha, K. K. Ghosh, S. Bhowmik, and R. Sarkar, "Mutually Informed Correlation Coefficient (MICC) – a New Filter Based Feature Selection Method," in *IEEE CALCON*, 2020.

[15] M. Ghosh, S. Begum, R. Sarkar, D. Chakraborty, and U. Maulik, "Recursive Memetic Algorithm for gene selection in microarray data," *Expert Syst. Appl.*, vol. 116, pp. 172–185, 2019.

[16] M. Ghosh, S. Adhikary, K. K. Ghosh, A. Sardar, S. Begum, and R. Sarkar, "Genetic algorithm based cancerous gene identification from microarray data using ensemble of filter methods," *Med. Biol. Eng. Comput.*, vol. 57, no. 1, pp. 159–176, 2019.

[17] M. Ghosh, S. Malakar, S. Bhowmik, R. Sarkar, and M. Nasipuri, "Feature Selection for Handwritten Word Recognition Using Memetic Algorithm," in *Advances in Intelligent Computing*, Springer, 2019, pp. 103–124.

[18] M. M. Kabir, M. Shahjahan, and K. Murase, "A new hybrid ant colony optimization algorithm for feature selection," *Expert Syst. Appl.*, vol. 39, no. 3, pp. 3747–3763, 2012.

[19] J. M. Cadenas, M. C. Garrido, and R. MartíNez, "Feature subset selection filter–wrapper based on low quality data," *Expert Syst. Appl.*, vol. 40, no. 16, pp. 6241–6252, 2013.

[20] M. Ghosh, R. Guha, R. Sarkar, and A. Abraham, "A wrapper-filter feature selection technique based on ant colony optimization," *Neural Comput. Appl.*, pp. 1–19, 2019.

[21] S. Mirjalili and A. Lewis, "The whale optimization algorithm," *Adv. Eng. Softw.*, vol. 95, pp. 51–67, 2016.

[22] M. M. Mafarja and S. Mirjalili, "Hybrid Whale Optimization Algorithm with simulated annealing for feature selection," *Neurocomputing*, vol. 260, pp. 302–312, 2017.

[23] M. Mafarja and S. Mirjalili, "Whale optimization approaches for wrapper feature selection," *Appl. Soft Comput.*, vol. 62, pp. 441–453, 2018.

[24] M. Sharawi, H. M. Zawbaa, and E. Emary, "Feature selection approach based on whale optimization algorithm," in *2017 Ninth International Conference on Advanced Computational Intelligence (ICACI)*, 2017, pp. 163–168.

[25] A. G. Hussien, A. E. Hassanien, E. H. Houssein, S. Bhattacharyya, and M. Amin, "S-shaped binary whale optimization algorithm for feature selection," in *Recent trends in signal and image processing*, Springer, 2019, pp. 79–87.

[26] A. G. Hussien, E. H. Houssein, and A. E. Hassanien, "A binary whale optimization algorithm with hyperbolic tangent fitness function for feature selection," in *2017 Eighth International Conference on Intelligent Computing and Information Systems (ICICIS)*, 2017, pp. 166–172.

[27] R. Eberhart and J. Kennedy, "A new optimizer using particle swarm theory," in *Micro Machine and Human Science, 1995. MHS'95., Proceedings of the Sixth International Symposium on*, 1995, pp. 39–43.

[28] F. van den Bergh and A. P. Engelbrecht, "A new locally convergent particle swarm optimiser," in *IEEE International conference on systems, man and cybernetics*, 2002, vol. 3, pp. 6-pp.

[29] E. Rashedi, H. Nezamabadi-pour, and S. Saryazdi, "GSA: A Gravitational Search Algorithm," *Inf. Sci. (Ny).*, vol. 179, no. 13, pp. 2232–2248, 2009.

[30] S. Mirjalili and S. Z. M. Hashim, "A new hybrid PSOGSA algorithm for function optimization," *Proc. ICCIA 2010 - 2010 Int. Conf. Comput. Inf. Appl.*, no. 1, pp. 374–377, 2010.

[31] M. Ghosh, R. Guha, I. Alam, P. Lohariwal, D. Jalan, and R. Sarkar, "Binary Genetic Swarm Optimization: A Combination of GA and PSO for Feature Selection," *J. Intell. Syst.*, vol. 29, no. 1, pp. 1598–1610, 2019.

[32] M. E. Basiri and S. Nemati, "A Novel Hybrid ACO-GA Algorithm for Text Feature Selection," pp. 2561–2568, 2009.

[33] X. H. Shi, Y. C. Liang, H. P. Lee, C. Lu, and L. M. Wang, "An improved GA and a novel PSO-GA-based hybrid algorithm," *Inf. Process. Lett.*, vol. 93, no. 5, pp. 255–261, 2005.

[34] M. Ghosh, R. Guha, P. K. Singh, V. Bhateja, and R. Sarkar, "A histogram based fuzzy ensemble technique for feature selection," *Evol. Intell.*, pp. 1–12, 2019.

[35] R. Guha *et al.*, "Deluge based Genetic Algorithm for feature selection," *Evol. Intell.*, pp. 1–11, 2019.

[36] L.-Y. Chuang, H.-W. Chang, C.-J. Tu, and C.-H. Yang, "Improved binary PSO for feature selection using gene expression data," *Comput. Biol. Chem.*, vol. 32, no. 1, pp. 29–38, 2008.

[37] S. M. Vieira, L. F. Mendonça, G. J. Farinha, and J. M. C. Sousa, "Modified binary PSO for feature selection using SVM applied to mortality prediction of septic patients," *Appl. Soft Comput.*, vol. 13, no. 8, pp. 3494–3504, 2013.

[38] B. Xue, M. Zhang, and W. N. Browne, "Multi-objective particle swarm optimisation (PSO) for feature selection," in *Proceedings of the 14th annual conference on Genetic and evolutionary computation*, 2012, pp. 81–88.

[39] E. Emary, H. M. Zawbaa, and A. E. Hassanien, "Binary ant lion approaches for feature selection," *Neurocomputing*, vol. 213, pp. 54–65, 2016.

[40] H. M. Zawbaa, E. Emary, and B. Parv, "Feature selection based on antlion optimization algorithm," in *2015 Third World Conference on Complex Systems (WCCS)*, 2015, pp. 1–7.

[41] I. Aljarah, H. Faris, and S. Mirjalili, "Optimizing connection weights in neural networks using the whale optimization algorithm," *Soft Comput.*, vol. 22, no. 1, pp. 1–15, 2018.

[42] D. Oliva, M. A. El Aziz, and A. E. Hassanien, "Parameter estimation of photovoltaic cells using an improved chaotic whale optimization algorithm," *Appl. Energy*, vol. 200, pp. 141–154, 2017.

[43] D. B. Prakash and C. Lakshminarayana, "Optimal siting of capacitors in radial distribution network using whale optimization algorithm," *Alexandria Eng. J.*, vol. 56, no. 4, pp. 499–509, 2017.

[44] A. Kaveh and M. I. Ghazaan, "Enhanced whale optimization algorithm for sizing optimization of skeletal structures," *Mech. Based Des. Struct. Mach.*, vol. 45, no. 3, pp. 345–362, 2017.

[45] J. Wang, P. Du, T. Niu, and W. Yang, "A novel hybrid system based on a new proposed algorithm—Multi-Objective Whale Optimization Algorithm for wind speed forecasting," *Appl. Energy*, vol. 208, pp. 344–360, 2017.

[46] H. M. Zawbaa, E. Emary, and C. Grosan, "Feature selection via chaotic antlion optimization," *PLoS One*, vol. 11, no. 3, p. e0150652, 2016.

[47] S. Ahmed, M. Mafarja, H. Faris, and I. Aljarah, "Feature selection using salp swarm algorithm with chaos," in *Proceedings of the 2nd International Conference on Intelligent Systems, Metaheuristics & Swarm Intelligence*, 2018, pp. 65–69.

[48] C.-S. Yang, L.-Y. Chuang, J.-C. Li, and C.-H. Yang, "Chaotic maps in binary particle swarm optimization for feature selection," in *2008 IEEE Conference on Soft Computing in Industrial Applications*, 2008, pp. 107–112.

[49] G. I. Sayed, A. Tharwat, and A. E. Hassanien, "Chaotic dragonfly algorithm: an improved metaheuristic algorithm for feature selection," *Appl. Intell.*, vol. 49, no. 1, pp. 188–205, 2019.

[50] G. I. Sayed, A. E. Hassanien, and A. T. Azar, "Feature selection via a novel chaotic crow search algorithm," *Neural Comput. Appl.*, vol. 31, no. 1, pp. 171–188, 2019.

[51] G. I. Sayed, A. Darwish, and A. E. Hassanien, "A new chaotic whale optimization algorithm for features selection," *J. Classif.*, vol. 35, no. 2, pp. 300–344, 2018.

[52] E. Tanyildizi and T. Cigal, "Continuous Time Chaotic Systems for Whale Optimization Algorithm," *Adv. Electr. Comput. Eng.*, vol. 18, no. 4, pp. 49–57, 2018.

[53] F. Mohanty, S. Rup, and B. Dash, "An Improved CAD Framework for Digital Mammogram Classification Using Compound Local Binary Pattern and Chaotic Whale Optimization-Based Kernel Extreme Learning Machine," in *International Conference on Artificial Neural Networks*, 2018, pp. 14–23.

[54] M. Wang and H. Chen, "Chaotic multi-swarm whale optimizer boosted support vector machine for medical diagnosis," *Appl. Soft Comput.*, vol. 88, p. 105946, 2020.

[55] D. Prasad, A. Mukherjee, G. Shankar, and V. Mukherjee, "Application of chaotic whale optimisation algorithm for transient stability constrained optimal power flow," *IET Sci. Meas. Technol.*, vol. 11, no. 8, pp. 1002–1013, 2017.

[56] C. J. Santana Jr, M. Macedo, H. Siqueira, A. Gokhale, and C. J. A. Bastos-Filho, "A novel binary artificial bee colony algorithm," *Futur. Gener. Comput. Syst.*, vol. 98, pp. 180–196, 2019.

[57] N. Kushwaha and M. Pant, "Link based BPSO for feature selection in big data text clustering," *Futur. Gener. Comput. Syst.*, vol. 82, pp. 190–199, 2018.

[58] S. Saha *et al.*, "Feature Selection for Facial Emotion Recognition Using Cosine Similarity-Based Harmony Search Algorithm," *Appl. Sci.*, vol. 10, no. 8, p. 2816, 2020.

[59] B. Chatterjee, T. Bhattacharyya, K. K. Ghosh, P. K. Singh, Z. W. Geem, and R. Sarkar, "Late Acceptance Hill Climbing Based Social Ski Driver Algorithm for Feature Selection," *IEEE Access*, vol. 8, pp. 75393–75408, 2020.

[60] I. Chatterjee, M. Ghosh, P. K. Singh, R. Sarkar, and M. Nasipuri, "A clustering-based feature selection framework for handwritten Indic script classification," *Expert Syst.*, vol. 36, no. 6, p. e12459, 2019.

[61] S. Sen, M. Mitra, A. Bhattacharyya, R. Sarkar, F. Schwenker, and K. Roy, "Feature selection for recognition of online handwritten bangla characters," *Neural Process. Lett.*, vol. 50, no. 3, pp. 2281–2304, 2019.

[62] K. K. Ghosh, S. Ahmed, P. K. Singh, Z. W. Geem, and R. Sarkar, "Improved Binary Sailfish Optimizer based on Adaptive β-Hill Climbing for Feature Selection," *IEEE Access*, p. 1, 2020.

[63] A. A. Heidari, S. Mirjalili, H. Faris, I. Aljarah, M. Mafarja, and H. Chen, "Harris hawks optimization: Algorithm and applications," *Futur. Gener. Comput. Syst.*, vol. 97, pp. 849–872, 2019.

[64] A. Fahad, Z. Tari, I. Khalil, A. Almalawi, and A. Y. Zomaya, "An optimal and stable feature selection approach for traffic classification based on multi-criterion fusion," *Futur. Gener. Comput. Syst.*, vol. 36, pp. 156–169, 2014.

[65] R. Chatterjee, T. Maitra, S. K. H. Islam, M. M. Hassan, A. Alamri, and G. Fortino, "A novel machine learning based feature selection for motor imagery EEG signal classification in Internet of medical things environment,"

| | |
|---|---|
| | *Futur. Gener. Comput. Syst.*, vol. 98, pp. 419–434, 2019. |
| [66] | G. Sun, J. Li, J. Dai, Z. Song, and F. Lang, "Feature selection for IoT based on maximal information coefficient," *Futur. Gener. Comput. Syst.*, vol. 89, pp. 606–616, 2018. |
| [67] | M. Ghosh, S. K. Bera, R. Guha, and R. Sarkar, "Contrast Enhancement of Degraded Document Image using Partitioning based Genetic Algorithm," in *International Conference on Emerging Technologies for Sustainable Development (ICETSD '19)*, 2019, pp. 431–435. |
| [68] | M. M. Mafarja and S. Mirjalili, "Hybrid binary ant lion optimizer with rough set and approximate entropy reducts for feature selection," *Soft Comput.*, vol. 23, no. 15, pp. 6249–6265, 2019. |
| [69] | Q. Al-Tashi, S. J. A. Kadir, H. M. Rais, S. Mirjalili, and H. Alhussian, "Binary Optimization Using Hybrid Grey Wolf Optimization for Feature Selection," *IEEE Access*, vol. 7, pp. 39496–39508, 2019. |
| [70] | S. Dhargupta, M. Ghosh, S. Mirjalili, and R. Sarkar, "Selective opposition based grey wolf optimization," *Expert Syst. Appl.*, p. 113389, 2020. |
| [71] | M. M. Kabir, M. Shahjahan, and K. Murase, "A new local search based hybrid genetic algorithm for feature selection," *Neurocomputing*, vol. 74, no. 17, pp. 2914–2928, 2011. |
| [72] | R. Guha, M. Ghosh, P. K. Singh, R. Sarkar, and M. Nasipuri, "M-HMOGA: A New Multi-Objective Feature Selection Algorithm for Handwritten Numeral Classification," *J. Intell. Syst.*, vol. 29, no. 1, pp. 1453–1467, 2019. |
| [73] | B. Alatas, E. Akin, and A. B. Ozer, "Chaos embedded particle swarm optimization algorithms," *Chaos, Solitons & Fractals*, vol. 40, no. 4, pp. 1715–1734, 2009. |
| [74] | B. Chen, W. Zeng, Y. Lin, and D. Zhang, "A new local search-based multiobjective optimization algorithm," *IEEE Trans. Evol. Comput.*, vol. 19, no. 1, pp. 50–73, 2015. |
| [75] | M. Ghosh, S. Adhikary, K. K. Ghosh, A. Sardar, S. Begum, and R. Sarkar, "Genetic algorithm based cancerous gene identification from microarray data using ensemble of filter methods," *Med. Biol. Eng. Comput.*, vol. 57, no. 1, pp. 159–176, 2019. |
| [76] | K. Priyanka and B. D. Kavita, "Feature selection using genetic algorithm and classification using weka for ovarian cancer," *Int. J. Comput. Sci. Inf. Technol.*, vol. 7, no. 1, pp. 194–196, 2016. |
| [77] | M. Ghosh, T. Kundu, D. Ghosh, and R. Sarkar, "Feature selection for facial emotion recognition using late hill-climbing based memetic algorithm," *Multimed. Tools Appl.*, vol. 78, no. 18, pp. 25753–25779, Jun. 2019. |